# Fusing Cross-Domain Knowledge from Multimodal Data to Solve Problems in the Physical World

YU ZHENG[1,2,3,4]
[1] JD Technology
[2] School of Computing and Artificial Intelligence, Southwest Jiaotong University
[3] School of Cyber Engineering, Xidian University
[4] Beijing Key Laboratory of Traffic Data Mining and Embodied Intelligence

The proliferation of artificial intelligence has enabled a diversity of applications that bridge the gap between digital and physical worlds. As physical environments are too complex to model through a single information acquisition approach, it is crucial to fuse multimodal data generated by different sources, such as sensors, devices, systems, and people, to solve a problem in the real world. Unfortunately, it is neither applicable nor sustainable to deploy new resources to collect original data from scratch for every problem. Thus, when data is inadequate in the domain of problem, it is vital to fuse knowledge from multimodal data that is already available in other domains. We call this cross-domain knowledge fusion. Existing research focus on fusing multimodal data in a single domain, supposing the knowledge from different datasets is intrinsically aligned; however, this assumption may not hold in the scenarios of cross-domain knowledge fusion. In this paper, we formally define the cross-domain multimodal data fusion problem, discussing its unique challenges, differences and advantages beyond data fusion in a single domain. We propose a four-layer framework, consisting of Domains, Links, Models and Data layers, answering three key questions: "what to fuse", "why can be fused", and "how to fuse". The Domains Layer selects relevant data from different domains for a given problem. The Links Layer reveals the philosophy of knowledge alignment beyond specific model structures. The Models Layer provides two knowledge fusion paradigms based on the fundamental mechanisms for processing data. The Data Layer turns data of different structures, resolutions, scales and distributions into a consistent representation that can be fed into an AI model. With this framework, we can design solutions that fuse cross-domain multimodal data effectively for solving real-world problems.

## 1. INTRODUCTION

The advances in sensing technology, large-scale computing infrastructures and artificial intelligence have fostered a variety of applications that solve real-world problems through

Authors' addresses: Y. Zheng, 13F, Block A, Building 2, No.20 Kechuang 11 Street, BDA, Beijing 100176, China; email: msyuzheng@outlook.com.

interactions with both virtual and physical worlds. As the natural processes and physical environment are very complex, it is rare that a single information acquisition method provides complete understanding of a phenomenon of interest [22]. Thus, a diversity of data obtained from different perspectives and sources using different types of instruments and measurement techniques (a.k.a. multimodal data) are used together to solve a problem [65][67]. Extensive studies have shown that fusing multimodal data achieves a better performance than using a single dataset in many tasks [1][15][32][40][63].

Recently, the big progress made in large language models [48][47][18] and embodied AI[26][30][52] has improved the capability of multimodal data fusion to some extent, attracting an even broader range of attentions to this topic from a diversity of communities [60][64]. For example, AI models can generate images or video clips based on text descriptions [8][39], or integrate videos, audios and text as an input to better understand human expression [44][47]. In these scenarios, we need to transfer the knowledge from one modality of data to another (i.e. cross-modalities) or complement knowledge from multiple data modalities to achieve a common goal (i.e. multi-modalities), such as understanding a phenomenon of interest or solving a problem.

However, most existing techniques only fuse multimodal data from a single domain mainly for solving problems in the digital world.

**1) Single Domain vs Multiple Domains:** For instance, as depicted in Figure 1 A), images, texts and videos from the same webpage would be fused to obtain a deeper understanding of a tourist attraction. Alternative, as shown in Figure 1 B), texts are used as an input to generate an image which will be gradually updated given more descriptions in texts. As illustrated in Figure 1 C), data generated by visual, audial and pressure sensors installed in a robot could be aggregated to better model the physical environment around the robot. In a brain study, as depicted in Figure 1 D), we would install multiple electric and magnetic sensors on different parts of a head mount device to capture signals of brains. In these scenarios, though multiple datasets of different modalities are fed into an AI model, they are originally created for the same purpose, e.g. describing the same location, modeling the same environment, and sensing the same head, in a single and specific domain, such as a webpage, an interface of a system, a robot, or a brain device. That is, the knowledge those datasets contain is intrinsically aligned at the very beginning of a task.

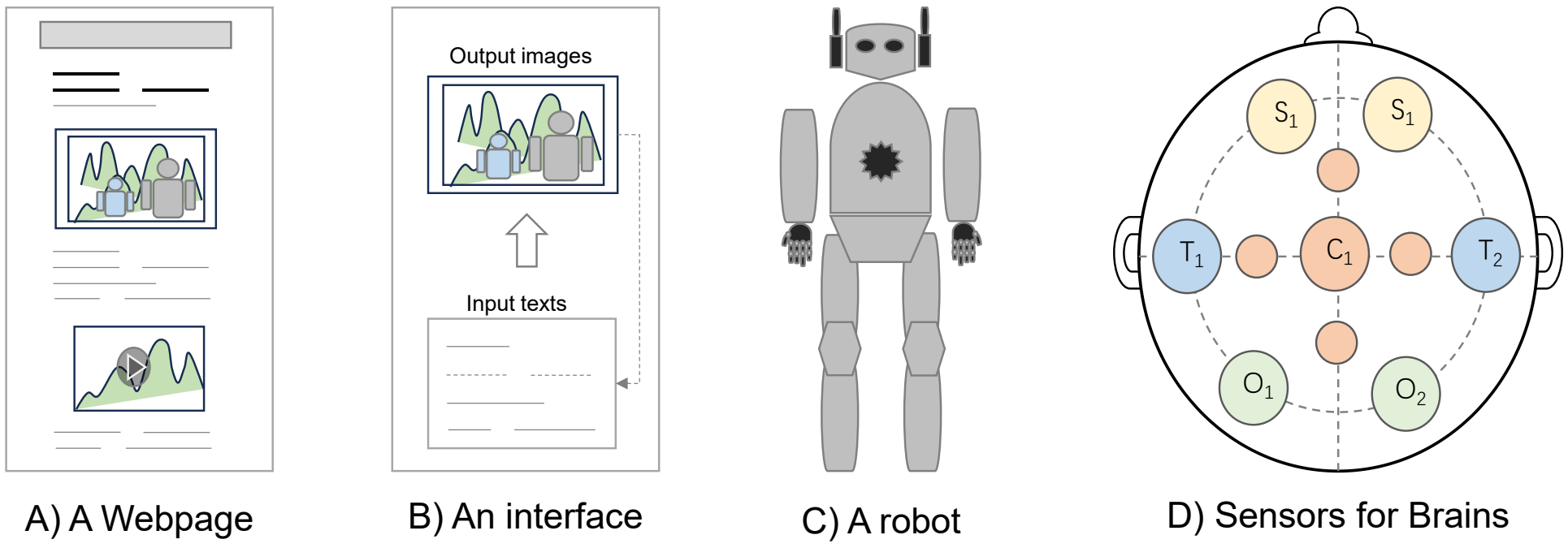

**Figure 1 Examples of data fusion in a single domain**

In the real world, however, there are more application scenarios in which we need to fuse multimodal data from different domains for many reasons, such as a limited budget, space constraints or time cost [67][68][69]. It is neither applicable nor sustainable to deploy new resources, including sensors, devices, instruments and labors, to collect original data from scratch for every task. Thus, to complete a task in a domain, as illustrated in Figure 2, we need to leverage additional datasets generated in other domains when the necessary data is insufficient in the domain of task. For example, in the domain of environment protection (denoted as Domain $\mathcal{A}$ in Figure 2), there is a problem of air pollution which would damage people's health. To solve this problem, we may design a few tasks, such as inferring the real-time and fine-grain air quality throughout an entire city ($T_1$) and then forecasting the air quality over the next 48 hours ($T_2$). To complete task $T_1$, we need to harness a diversity of datasets relating to traffic, land uses, and meteorology besides air quality data. This is a task about environmental protection (belonging to Domain $\mathcal{A}$), while the datasets fed into an AI model are obtained from multiple domains, including transportations, urban planning and climate (i.e. Domain $\mathcal{B}$, $\mathcal{C}$, and $\mathcal{D}$). As those datasets were not originally collected for inferring air quality, the knowledge from them is not naturally aligned.

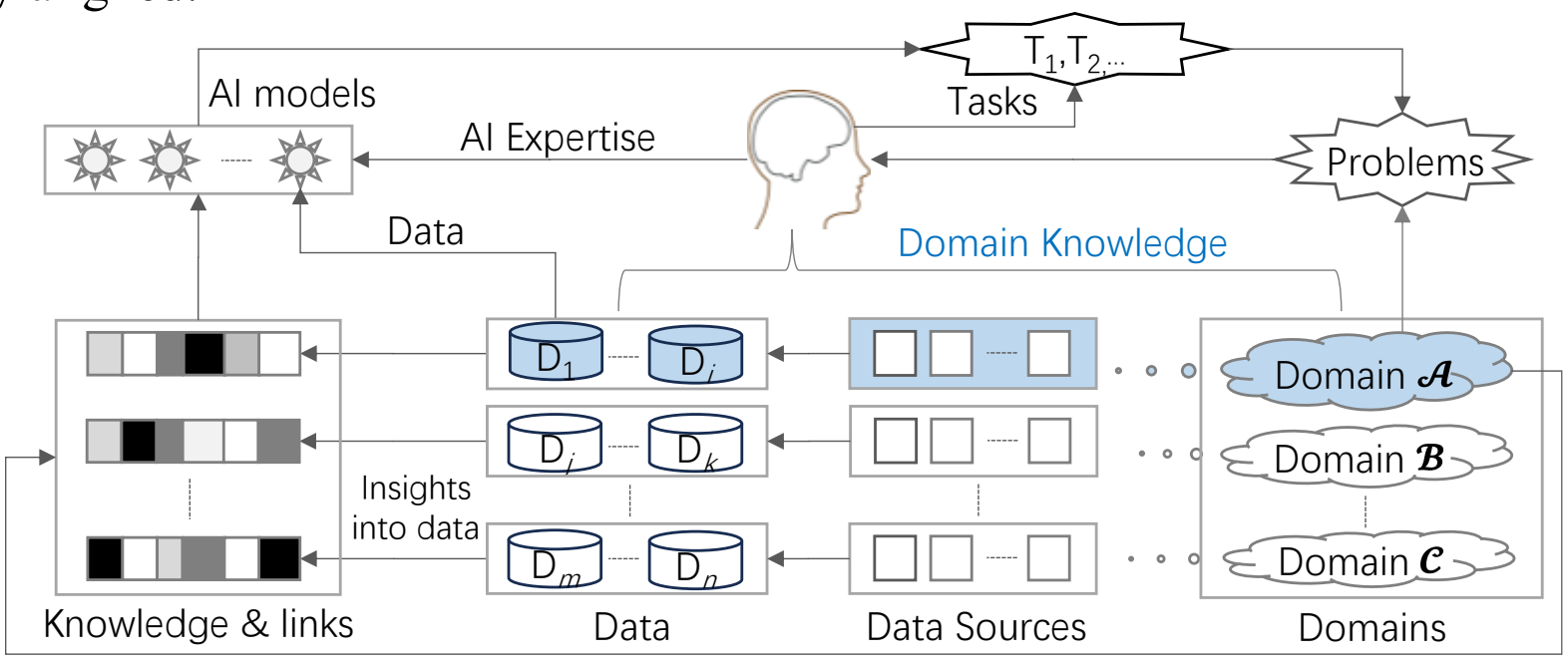

**Figure 2. Illustration of cross-domain knowledge fusion**

**2) Problems in Digital World vs in Physical World**: There are three scenarios of problem solving in which we fuse multimodal data, as illustrated in Figure 3. Most research has been done in the first scenario and a few are stepping into the second one. However, the last one is very rare, which is the main target of this paper, though our framework is applicable to all these three scenarios.

The first one fuses multimodal data obtained from the digital world in a task that is designed to solve problems also in the digital world, as depicted in Figure 3 A). For example, generating a video clip using an AI model based on text descriptions for demonstrating an idea.

The second one fuses multimodal data from both digital and physical worlds in a task for solving a problem in the digital world, as shown in Figure 3 B). For instance, integrating data obtained from handle or infrared light sensors deployed in the physical world for playing a Wii or Kinect video game in the digital world.

The third one fuses multimodal data from both worlds in a task running in digital world but for solving a problem in the physical world, as illustrated in Figure 3 C). For example, combing weather conditions, traffic data and social media to adjust traffic control policies in a city's downtown. As the cost of data collection in the physical world is much higher than in the digital world, it is not easy to have abundant data in every application scenario. In many cases, data scarcity and data missing problems are even more difficult to solve than handling data diversities. Under such a circumstance, combining domain knowledge with data becomes more important for solving a real problem.

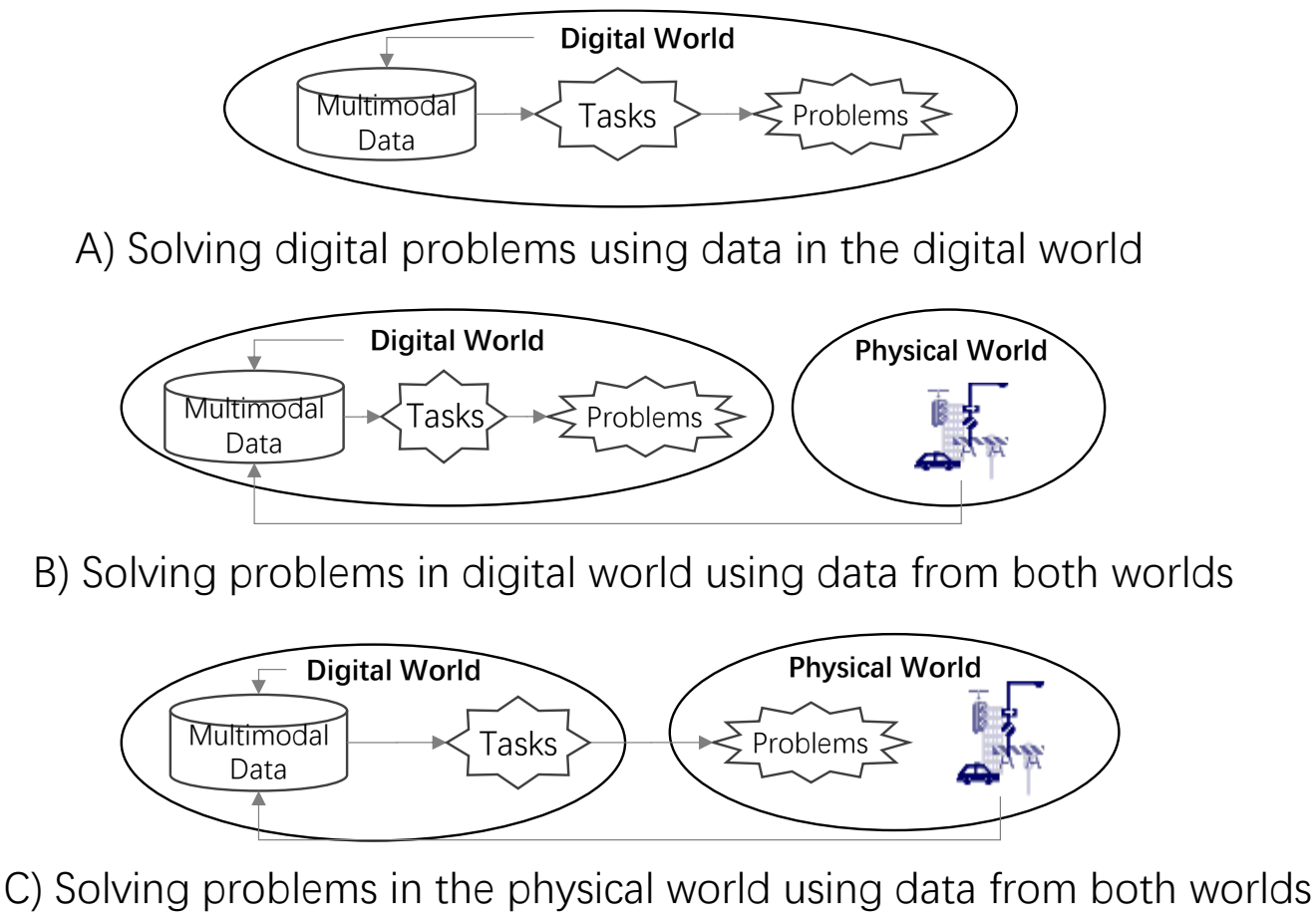

**Figure 3 Three problem solving scenarios when fusing multimodal data**

The above-mentioned two differences (single vs multiple domains, virtual vs physical worlds) pose new challenges to data fusion, including knowledge alignment across disparate data from different domains and handling data insufficiency in each dataset during fusion. As a result, we need to answer the following three questions before placing data into a model's structure.

1) What kind of datasets can be selected from other domains? i.e. ***what*** to fuse?
2) Why can these datasets be fused together to achieve a better performance? i.e. the underlying links between those datasets. In short, ***why*** can be fused?
3) How to design a specific model structure to fuse the knowledge from selected data? i.e. ***how*** to fuse?

In task $T_1$, the selection of data depends on domain knowledge from $\mathcal{A}$, $\mathcal{B}$, $\mathcal{C}$, and $\mathcal{D}$ and the insights into data, as shown in the bottom part of Figure 2. Regarding the second question, the links between data are derived from both domain knowledge and the philosophy of knowledge alignment. The answer to the third question depends on the link between datasets, the nature of data modalities and AI expertise. Based on the designed AI model, we can fuse the knowledge from these datasets to complete task $T_1$, and then solve the problem gradually.

In this paper, we formally define the cross-domain multimodal data fusion problem, exploiting its differences, challenges and advantages beyond data fusion in a single domain. This is a new and vital research theme for solving real-world problems. To answer the "*what*", "*why*" and "*how*" questions mentioned above, we propose a four-layer framework, consisting of the Domains, Links, Models and Data layers. The contributions of this article are four folds.

- In the Domain Layer, we propose a methodology to select relevant data from different domains for a given problem, through a procedure: problems → root causes → factors→data→sources and domains, i.e. answering the question of *what*.
- In the Links Layer, we reveal the philosophy of knowledge alignment beyond specific model structures, supporting the discovery of links between data. The philosophy consists of Multiview-based, Similarity-based, Dependency-based and Commonality-based principles, explaining the rationale of the complementation between knowledges from different data and thus answering the question of "why can be fused".

- In the Models Layer, we represent existing data fusion algorithms with two knowledge fusion paradigms, which is comprised of a precise fusion and a coarse fusion, based on their fundamental mechanisms for processing data. The two paradigms do not only explain the evolution of data fusion algorithms but also unveil the intrinsic differences between existing methods, supporting the design of a specific AI model for fusing multimodal data, i.e. answering the question of "how can be fused".
- In the Data Layer, the proposed data transformation component turns data of different structures, resolutions, scales and distributions into a consistent representation that can be fed into an AI model. It consists of data preprocessing, precise transformation and coarse transformation, generating three data transformation approaches based on intrinsic properties of data modality and features of application scenarios.

## 2. PROBLEM DEFINITION

### 2.1 Preliminaries and Examples

This section formally defines a few terms, consisting of domains, data sources, data modalities, knowledge from data, AI tasks and the relationship among them, as illustrated in Figure 4.

**Domains and data sources**: A domain is an area of interest, such as transportation, environmental protection, public safety, economy, climate, and entertainment. The scope of a domain depends on the granularities of an interest, which is usually represented by a taxonomy. For example, the domain of environmental protection can be comprised of several sub-domains, including air quality, water quality, noise and soils, etc. A domain contains many data sources, which could be a sensor, a device, an instrument, a system or a person, constantly generating data.

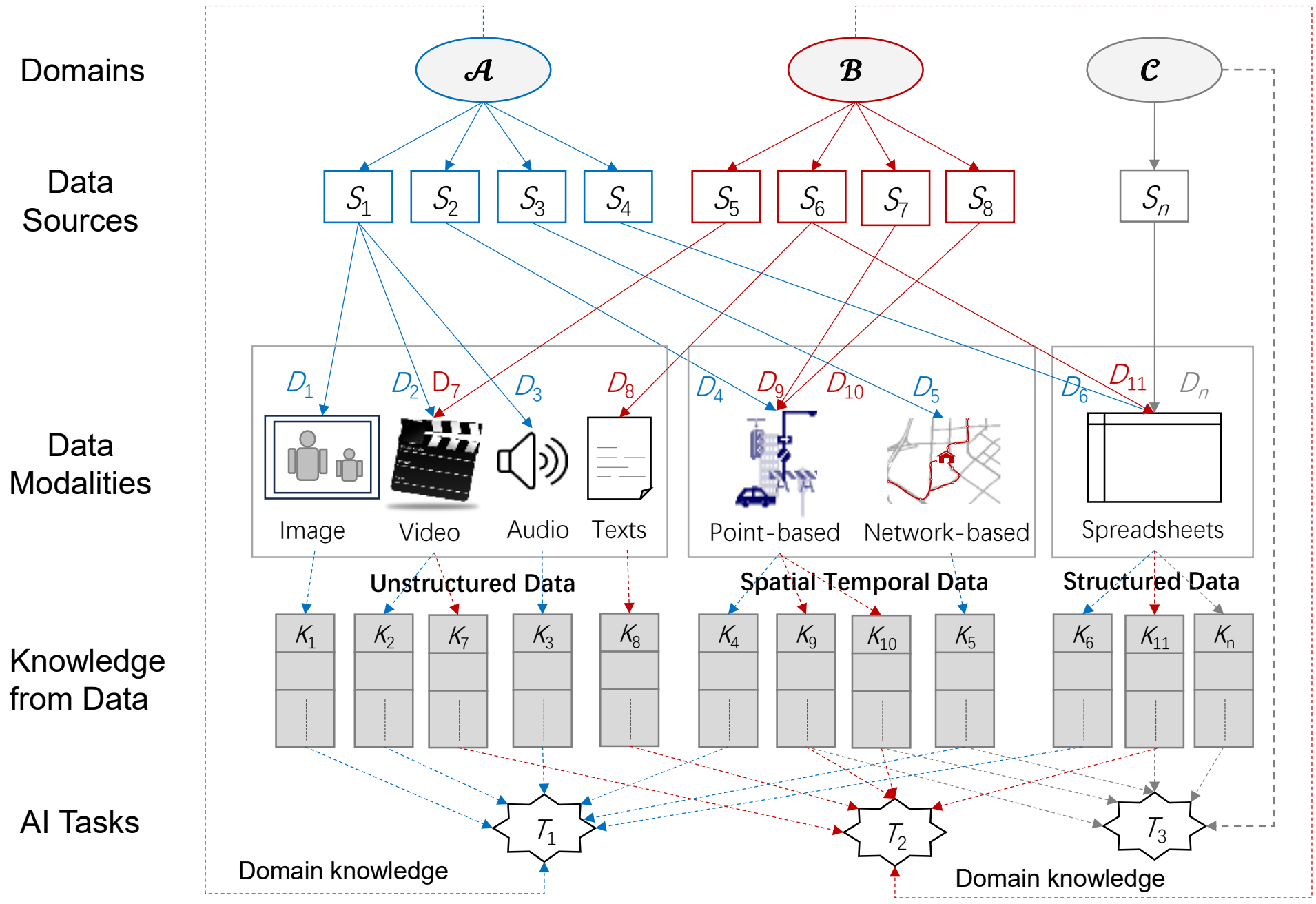

**Figure 4 Illustrations of key concepts in cross-domain multimodal data fusion**

**Data modalities**: The data generated by a source could have different modalities represented by different forms of data structures. In the real world, there are three categories of data modalities, consisting of unstructured data, spatio-temporal (ST) data and structured data. Images, audios, videos and texts are unstructured data. Representatives of ST data consist of point-based ST data, like IOT (Internet of Things) data, and network-based ST data, e.g. traffic networks and trajectories etc. Typical examples of structed data are digital spreadsheets generated in E-Government services.

**Knowledge**: Each piece of data contains a certain knowledge about the source and domain in which it was created. The knowledge has a diversity of representations denoted by hand-crafted features or model-generated vectors and matrices in latent spaces.

**AI Tasks**: Knowledges from different data can be fused into an AI model to complete a task, which is designed to solve a problem, based on the understanding of the problem and the relating domains. We call such kind of tasks an AI task in this paper.

**Cross-domain knowledge fusion** uses knowledge extracted from datasets generated in different domains (including a target domain and other domains) to solve a problem in the target domain. These datasets could have the same or different data modalities.

**Multimodal data fusion** denotes fusing knowledge from multiple datasets of different modalities to complete a task. These datasets could be generated a single or different domains. This is different from c**ross-modality knowledge fusion**, which aims to transfer knowledge from one data modality (e.g. texts) to another data modality (e.g. videos).

**Physical world problems**: In such a problem, a significant portion of datasets are collected in the physical world and are computed to understand the physical environment and natural processes. Computing results may be further applied to physical objects, such as sensors, devices, instruments, vehicles and buildings, for facilitating the operation of the real world.

Now, we can see (single vs cross) **domains**, (unimodal vs multimodal vs cross-modal) **data**, and (virtual and physical) **worlds** are three different dimensions defining a knowledge fusing task. Figure 4 presents a concrete example of cross-domain knowledge fusion from multimodal data in the physical world.

In the transportation domain ($\boldsymbol{\mathcal{A}}$), to manage a city's traffic, we need to create a few data sources, e.g. $S_1$, $S_2$, $S_3$ and $S_4$ depicted in Figure 4, from which a bunch of data ($D_1$~$D_6$) can be collected. Here, $S_1$ is a set of cameras sensing traffic conditions on roads. $S_1$ generates images capturing traffic violations of vehicle ($D_1$), videos about traffic density ($D_2$), and audios recording traffic noises ($D_3$). $S_2$ is a set of loop detectors deployed on roads, counting the number of vehicles traversing a road. It constantly creates point-based ST data ($D_4$). $S_3$ is a taxi dispatching system, which stores the GPS trajectories of taxicabs traveling in a city ($D_5$). $S_4$ is a system recording the penalties of traffic violation that have been processed by police officers ($D_6$). The knowledge $K_1$ extracted from $D_1$, $K_2$ from $D_2$, $K_3$ from $D_3$, $K_4$ from $D_4$, $K_5$ from $D_5$, and $K_6$ from $D_6$ are then fused in an AI model to complete task $T_1$, which aims to predict the impact area of a given traffic accident. This is a case of multimodal data fusion in a single domain.

In the environmental protection domain ($\boldsymbol{\mathcal{B}}$), to tackle the challenges of air pollution, we deploy four data sources, consisting of $S_5$, $S_6$, $S_7$ and $S_8$. $S_5$ is a set of cameras that monitor the pollution situation of a territory, generating video clips denoted as $D_7$. $S_6$ is a system that records penalties of pollution violation issued by governments, generating descriptions of violation ($D_8$) and records of penalty ($D_{11}$). There are two other data sources $S_7$ and $S_8$, which are two sets of sensors detecting the concentration of air pollutants in a city and the pollution emission from major factories respectively. They generate a sequence of point-based sensory readings, denoted as $D_9$ and $D_{10}$. The knowledge, including $K_7$ from $D_7$, $K_8$ from $D_8$, $K_9$ from $D_9$ etc., is then fed into a model to complete task $T_2$, which aims to forecast the air quality over the next 48 hours.

Both $T_1$ and $T_2$ belong to multimodal data fusion in a single domain, though different sources are employed. Now, in the domain of urban planning ($\boldsymbol{\mathcal{C}}$), there is another task $T_3$, which attempts to infer the function of a region. As a region is usually a mixture of different functions, such as business, entertainment, education and residential areas, which is an accumulation of human behaviors and urban development over a long period of time [58], it needs to fuse multiple data of different modalities. However, the needed data is neither fully available nor enough in the urban planning domain, which could only have householder survey data like $D_n$. It is impossible to deploy new devices and systems for collecting necessary data either. The most effective approach to solving this problem is to leverage the related data that has already been generated in domain $\boldsymbol{\mathcal{A}}$ and $\boldsymbol{\mathcal{B}}$.

Based on the domain knowledge of $\boldsymbol{\mathcal{A}}$, we know datasets $D_1$~$D_5$ imply traffic patterns of an area. Furthermore, $D_5$ contains knowledge about human mobility patterns in a region, as GPS trajectory data of taxicabs has pickup and drop-off points of each trip. Based on the domain knowledge of $\boldsymbol{\mathcal{B}}$, we know datasets $D_9$ and $D_{10}$ could denote the natural environment of a place. According to the domain knowledge of $\boldsymbol{\mathcal{C}}$, traffic patterns and nature environment of a region are key observations of the region's underlying functions [59]. As a result, we can use the knowledge from $D_5$ in domain $\boldsymbol{\mathcal{A}}$, the knowledge from $D_9$ and $D_{10}$ in domain $\boldsymbol{\mathcal{B}}$ and the knowledge from $D_{11}$ to complete task $T_3$ in domain $\boldsymbol{\mathcal{C}}$. This is a typical scenario of cross-domain knowledge fusion from multimodal data.

Table 1 shows the differences of different knowledge fusion problems in Domains, Data and Worlds dimensions. The last row defines the problem discussed in this paper.

**Table 1 Key differences of knowledge fusion problems**

| | **Domains** | | **Data** | | | **Worlds** | |
|---|---|---|---|---|---|---|---|
| | Single domain | Cross-domain | Unimodal | Multimodal | Cross-modality | Virtual | Physical |
| Fig. 1 A) | √ | | | √ | | √ | |
| Fig. 1 B) | √ | | | | √ | √ | |
| Fig. 1 C) and D) | √ | | | √ | | | √ |
| $T_1$,$T_2$ in Fig. 4 | √ | | | √ | | | √ |
| $T_3$ in Fig. 4 **ours** | | **√** | | **√** | | | **√** |

## 2.2 Differences between this Topic and Existing Research

Though there are already quite a few articles studying the problem of multimodal data fusion, the differences between our paper and these articles are three-folds:

**1) Different scopes of the problem**: Existing research fuse multimodal data generated in a single domain, while we fuse multimodal data across multiple different domains where diversities and modalities of data are more complex to handle. Besides videos, images, audios and texts, there are a diversity of spatio-temporal data, such as sensory data, POIs and road networks, and structured data like records from online forms that could be fused in a task. Moreover, the goals of data fusion include not only completing virtual tasks, like text and video generation, self-programming and gamming, in digital worlds but also solving real-world problems, such as traffic controls, disaster responses and autonomous driving in the physical environment. There are more challenges posed to data fusion in the physical world. On the one hand, all most all tasks are facing data scarcity problem, as the cost of data collection in the physical world is much higher than the virtual world. Thus, they may not be simply handled by deep-learning-based methods which learns complex relationships between data through hidden layer-based black boxes given abundant data. One the other hand, the problems we need to solve in the physical world usually require a more accurate, efficient and reliable AI models than in the virtual one.

**2) Different focuses of the problem**: Existing research focus on specific data processing techniques that transform multimodal datasets into different representations which can be further aggregated in a machine learning model, supposing these datasets are correlated and complementary to each other. That is, they do not necessarily study "what to fuse" and "why can be fused". However, in this paper, before stepping into a model, we focus on first selecting multimodal data relating to a problem from different sources. This calls for rich knowledge of domains and deep insights into the data generated in these domains. Second, we find the meaningful and specific links between selected data, which calls for a systematical and deep understanding of knowledge alignment. These two parts are rarely mentioned in existing articles.

**3) Different depths of the problem**: Existing research study multimodal data fusion techniques given a specific category of AI models, such as tensor decomposition [22] and deep neural networks [15][63][64]. They usually exploit how different datasets can be properly placed in the specific structure of a model so that the knowledge from these datasets can be aggregated to complete a task. For example, early fusion, intermediate fusion and later fusion are three approaches in deep-learning-based methods [63]. Similarly, Gao et al. [15] classify these methods into three categories, consisting of signal-level, feature-level and decision-level fusions. Alternatively, Zhang et al [64] divide data fusion methods into different categories, e.g. encoder-decoder-based methods, attention-based methods, graph neural network-based methods, and Generative neural network-based methods, based on model structures. However, in the real world, an AI model will not automatically jump out when we face a problem. Given taxonomies proposed in those survey papers, professionals are still puzzled in selecting a proper model for their problems. Model selection is an important step towards solving a problem, depending on features of the problem, functions of models, and the underlying principle of knowledge alignment. The principle is beyond model structures and across models, applicable to different scenarios and domains for selecting models, designing model structures and guiding the way from data integration to knowledge fusion. These require a much deeper thinking towards the nature of knowledge alignment than just connecting data in a given model.

## 3. MOTIVATIONS OF THE RESEARCH

### 3.1 Benefits of Cross-Domain Knowledge Fusion

The benefits of cross-domain knowledge fusion are two folds at least.

One is reducing the effort for data collection tremendously. We always find that data in a domain is not enough to solve a given problem because the physical world is very complex. It is neither feasible nor sustainable to deploy new resources for every and each AI task. Without an efficient approach to obtaining necessary data, the cost of data collection in an AI task will become the bottleneck preventing AI from being widely used.

The other is to improve the capability of solving a problem, achieving a more accurate forecast, an earlier detection of anomalies, and a more reliable estimation of distributions.

- **More accurate forecasts**: Extensive studies have shown that cross-domain knowledge fusion offers a more accurate forecast than using multimodal data from a single domain in many scenarios. For example, by combing air quality data, meteorological data, traffic data and points of interest (POIs) data from different domains, we can predict future air quality with a much higher accuracy than only using air quality and meteorological data from the environment protection domain [56][70].

- **Earlier detection of anomalies**: Cross-domain knowledge fusion can detect sophisticated anomalies that cannot be recognized by using a single dataset or multimodal data from a single domain, or detect an anomaly much earlier than solely using one dataset. As introduced in paper [72], an unusual event has just happened at location $r_1$, affecting its surrounding locations. As a result, the traffic flow entering $r_1$ from its surrounding locations increases 10 percent. Meanwhile, social media posts and bike rental flow around these locations change slightly. The deviation in each single dataset against its common pattern is not significant enough to be considered anomalous. However, when putting them together, we might be able to identify the anomaly, as the three datasets barely change simultaneously to that extent.
- **More reliable estimations**: Cross-domain knowledge fusion can lead to a more reliable estimation on the distribution of data, particularly when the data is very sparse [42][50]. For instance, [42] aims to estimate the distribution of traffic volume on each and every road segment throughout a city based on GPS trajectories of taxicabs. However, the flow of taxicabs is only a small portion of the entire traffic flow in a city. In addition, the trajectories generated by taxicabs are sparsely distributed on road segments, i.e. there is few trajectory data on many road segments. Thus, it is very hard to obtain a reliable estimation on road-segment-level traffic volume if we only use taxi trajectories. By fusing the knowledge about structures of a road segment (such as lengths, widths, indegrees, outdegrees, and the number of lanes) and POIs around the segment into the traffic volume estimation through a coupled matrix factorization model, we can tackle the challenges posed by the data sparsity and therefore achieve a much better performance beyond the methods solely using traffic data.

## 3.2 Challenges of Cross-Domain Multimodal Data Fusion

The main challenges of this problem include the following three aspects. Other challenges, such as data missing, data redundancy, and data noise, that also exist in single domain data fusion are not discussed here again.

**1) Selecting useful data**: It is a challenge to select useful datasets from disparate data sources to solve a given problem, depending on a thorough understanding of the problem and a deep insight into the data from other domains. This calls for rich knowledge about the domain of problem and the domains where data is leveraged.

**2) Designing links between data**: Even if relevant datasets are selected from different domains, it is still challenging to design the specific links (or interactions) between selected data. This needs to bridge the gap between two parts. One is not only the knowledge about the factors causing a problem but also a deep understanding of the interactions between these factors. The other is AI expertise, particularly the fundamental principle of knowledge alignment that is beyond and across different AI models. Additionally, the conflict between data scarcity and the complexity of problems in the physical world compromises the capability of automated machine learning techniques [55], which was designed to automatically learn relationships between data and labels. It remains a challenge in finding a balance between totally taking human out of learning applications and a fully hand-crafted AI design.

**3) Representation learning**: Different datasets have different forms of representation, distributions, scales and resolutions. It is a challenge to turn them into sharable and computable representations while preserving their original knowledge. Advanced encoding (or embedding) techniques are still missing for spatio-temporal data and structured data.

The representation learning algorithms may be different for different data modalities in different categories of application scenarios.

The first two challenges are rarely discussed in existing research, and the second one is the most challenging part as it needs to bridge the gap between domain knowledge and AI expertise. Most existing articles focus on tackling the third challenge. Additionally, most of them apply the representation learning techniques, which were originally designed for unstructured data, to structure data and spatio-temporal data. As a results, some key properties, such as spatial distances and hierarchies, are not preserved in the encoded results, therefore losing important knowledge after the representation learning.

## 4. GENERAL FRAMEWORK OF CROSS-DOMAIN KNOWLEDGE FUSION

Figure 5 presents the framework of the methodology for cross-domain knowledge fusion, which is comprised of four layers: Domains, Bridges, Models and Data Layers. The Domains Layer selects useful datasets for solving a problem, tackling the first challenge mentioned in Section 3.2. It answers the question of what to fuse. The Links Layer identifies the links between data for knowledge alignment, tackling the second challenge of cross-domain knowledge fusion. It explains why data can be fused. The Models Layer designs specific model structures to complete a task. The Data Layer transforms multimodal data into consistent representations, tackling the third challenges mentioned in Section 3.2. The Models and Data Layers answer the question of how to fuse.

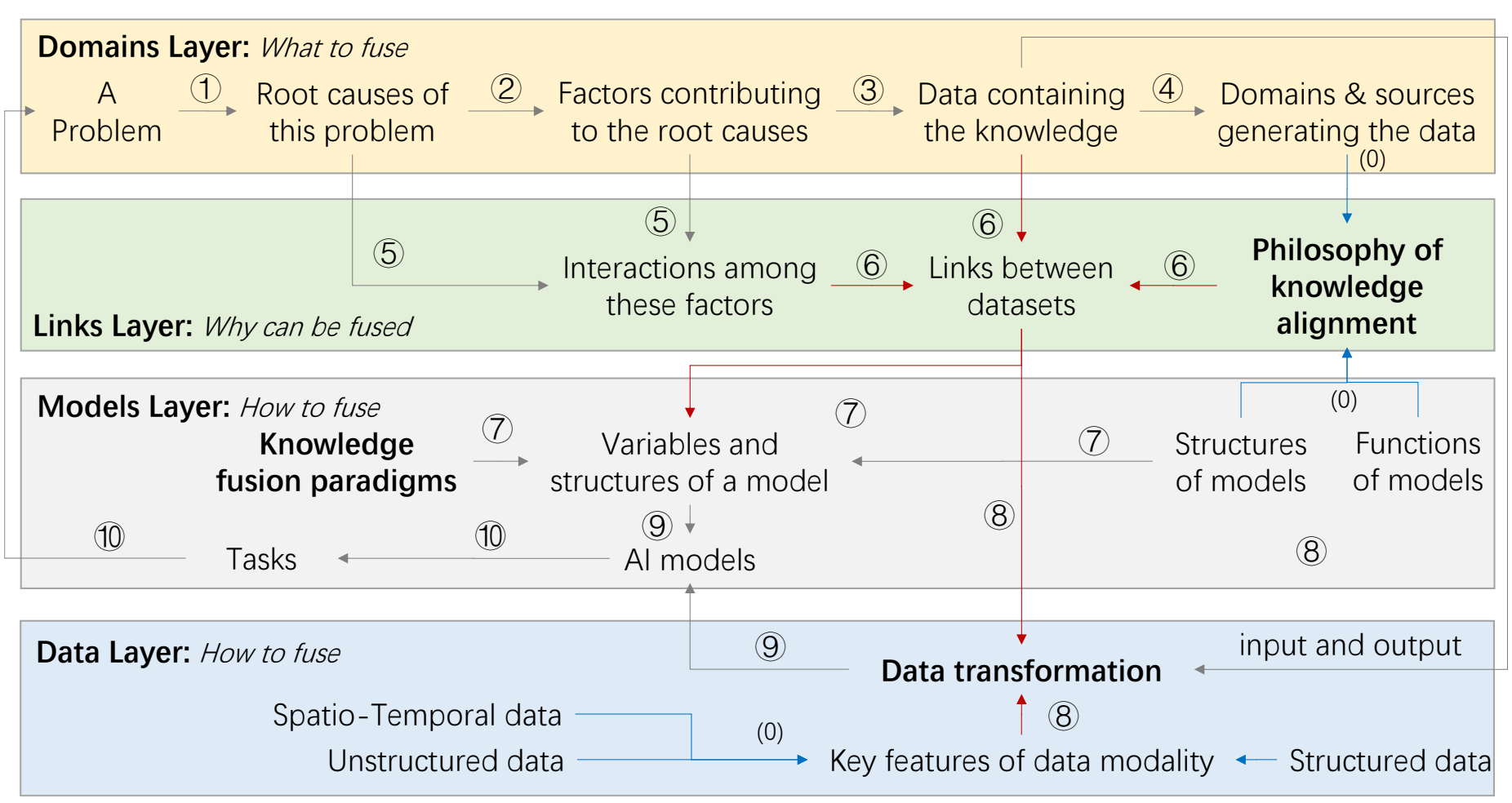

**Figure 5 Framework of the methodology for cross-domain multimodal data fusion**

### 4.1 Procedures

*In the Domains Layer:*

① **Analyze the root causes of the problem based on related domain knowledge.**
② **Digging out main factors contributing to the root causes**.
③ **Exploring relevant data containing the knowledge about these factors**.
④ **Searching for data sources and domains generating these datasets**.

*In the Links Layer:*

⑤ **Identifying the interactions between causal factors**. This step conceives coarse interactions between causal factors based on domain knowledge, as it is difficult to obtain precise interactions.

⑥ **Design the links between data based on three inputs**: the interactions between causal factors, the knowledge data contains, and the philosophy of knowledge alignment. The third one is the foundation deriving data fusion models, consisting of the Multiview-based, similarity-based, dependency-based, and commonality-based knowledge alignment principles (introduced in Section 5).

*In the Models & Data Layers:*

⑦ **Design an AI model with a specific structure and a set of variables, based on the links between selected data, the paradigm of knowledge fusion, and AI expertise.** There are two paradigms for knowledge fusion, consisting of precise fusion and coarse fusion, which will be introduced in Section 6. The coarse knowledge fusion paradigm only designs a high-level structure of an AI model and learns details from data during training processes.

⑧ **Design the data transformation algorithms for selected data based on the links between data and the key properties of different data modalities**. Different data modalities should have different transformation algorithms. In the meantime, the data of the same modality could have different data transformation algorithms in different application scenarios.

⑨ **The structures, variables and data transformation algorithms are coupled to construct final AI models**. Sometimes, the data transformation algorithms, e.g. deep encoders, is a part of a model's structure. The AI models are trained based on selected data using machine learning algorithms.

⑩ **Apply the designed AI models to the defined task to solve the problem**.

Figure 6 presents a summary of the above-mentioned procedures. It highlights three key components: the philosophy of knowledge alignment, knowledge fusion paradigms, and data transformation, which will be detailed in Section 5, 6 and 7 respectively. Most existing research focuses on the last two steps in this framework, i.e. design model structures and transform data as input, while ignoring the previous steps. In addition, existing data transformation methods mainly focus on processing unstructured data using embedding algorithms and encoders, lack of a systematic framework that can handle all types of data modalities for downstream knowledge fusion tasks in the physical world.

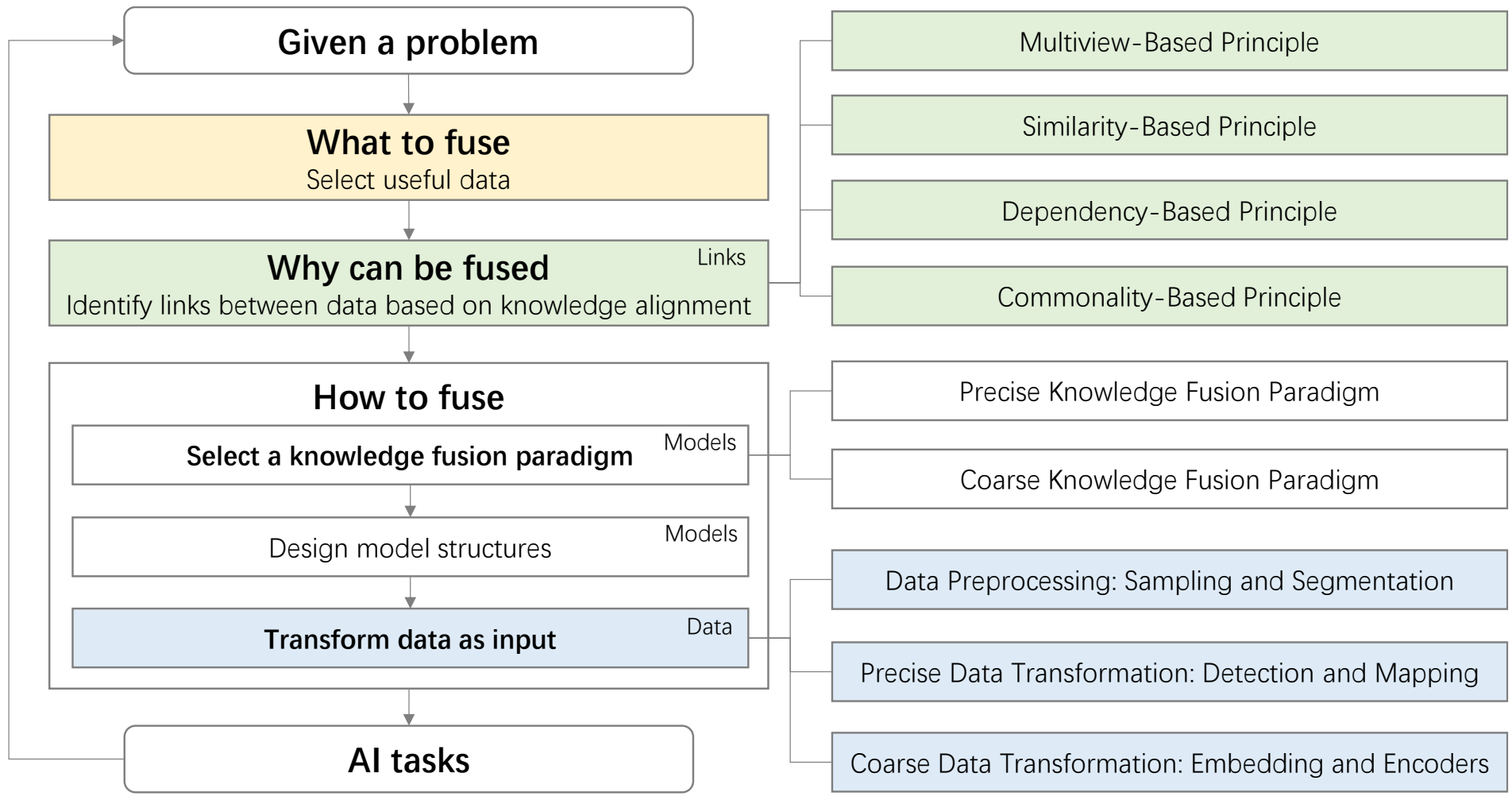

**Figure 6 Summary of cross-domain knowledge fusion from multimodal data**

More specifically, the philosophy of knowledge alignment has four main principles, each of which can enable a variety of AI models. There are two knowledge alignment paradigms, in which data are fused based on different processing and connecting mechanisms. The data transformation step depends on three main components, consisting of data preprocessing, precise transformation and coarse transformation, generating three data transformation approaches: precise, coarse, and hybrid transformation.

### 4.2 Example of Data Selection

As the first four procedures are application-driven and domain-related, we elaborate on them with a running example shown in Figure 7, which infers the real-time and fine-grained air quality throughout an entire city given the air quality readings from existing stations.

① Analyze the root causes: The air pollution in location A is mainly caused by three aspects: pollution emissions from location A and A's surrounding places, dispersion conditions of location A and A's surrounding places, and the secondary chemical reaction between different air pollutants in location A.

② Digging out main factors: Traffic conditions is a key factor contributing to traffic emission which is an important source of air pollution. Land uses (e.g. the density and height of buildings) and meteorology in a location contribute to the location's dispersion conditions. Meteorology is also a factor contributing to the secondary chemical reaction between air pollutants.

③ Exploring relevant data: POIs and road network data contain the knowledge about a location's land use as well as traffic conditions. GPS trajectories of vehicles travelling in a location contain the knowledge about traffic conditions in the location. Wind speeds, humidity and weather of a location denote its meteorological conditions.

④ Searching for data sources: we can obtain GPS trajectories of taxicabs, which is a portion of traffic flow on roads, from taxi dispatching companies. POIs and network data can be obtained from a map service provider in transportation domain. Meteorological data can be collected from a diversity of sensors through the information system deployed in the bureau of meteorology.

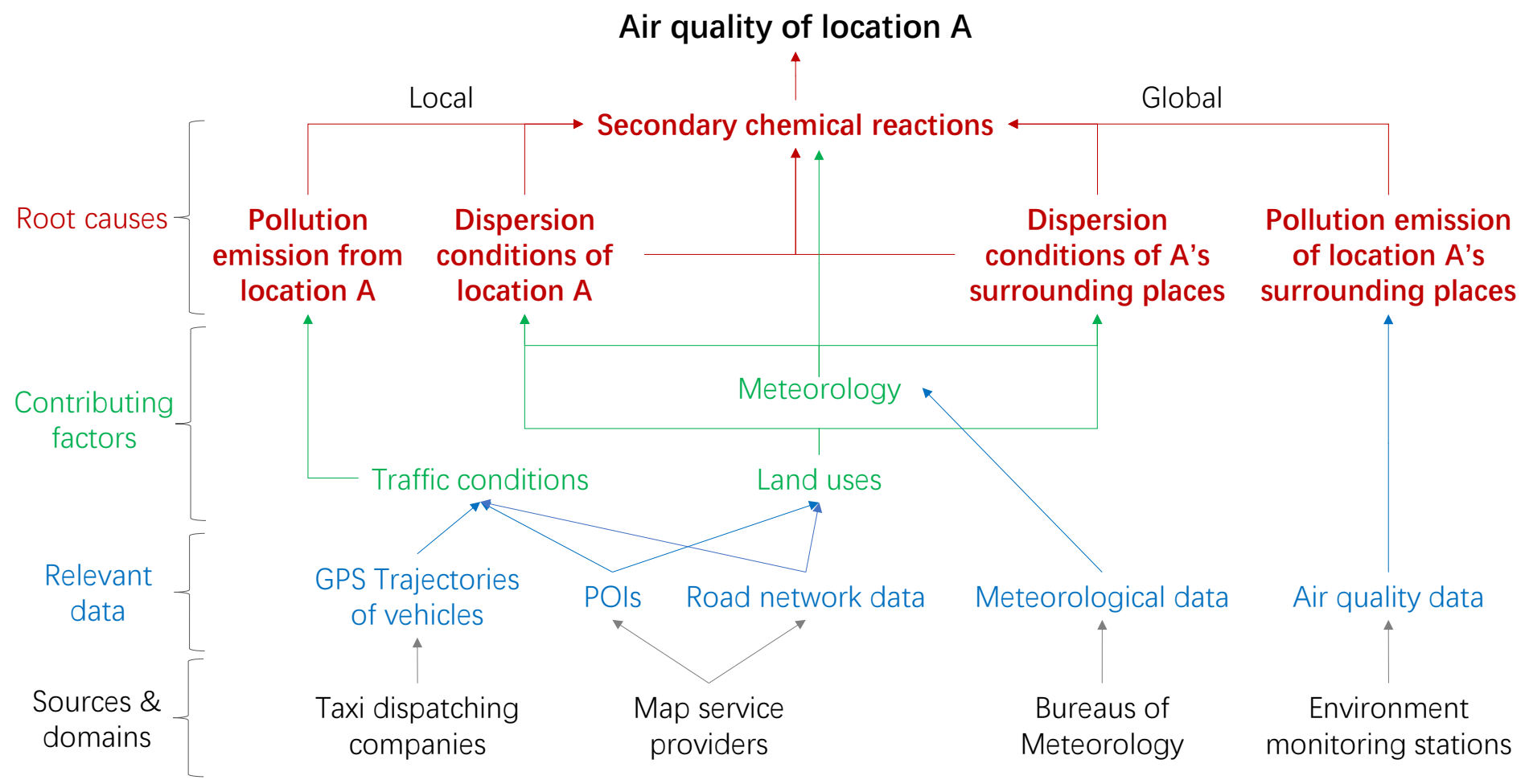

**Figure 7 Example of selecting relevant data for a given problem**

## 5. PHILOSOPHY OF KNOWLEDGE ALIGNMENT

This section presents the philosophy of knowledge alignment, which reveals the nature of complementation between knowledge from different data, regardless structures of AI models. The philosophy provides four knowledge alignment principles, as illustrated in Figure 8, supporting the design of cross-domain knowledge fusion models. Each principle can foster a wide range of AI model structures, including both deep learning-based models like Convolutional Neural Networks (CNN), Long-Short Term Memory (LSTM), Graph Convolutional Networks (GCN) and generative pre-trained Transformer (GPT), and non-deep learning-based models, such as coupled matrix factorization, co-training and probabilistic graphical models.

There are four labels in Figure 8, denoting data, latent representations, objects and domains respectively. Here, we only use two objects and two domains for a simple illustration. The numbers can be greater than 2 in the philosophy of knowledge alignment.

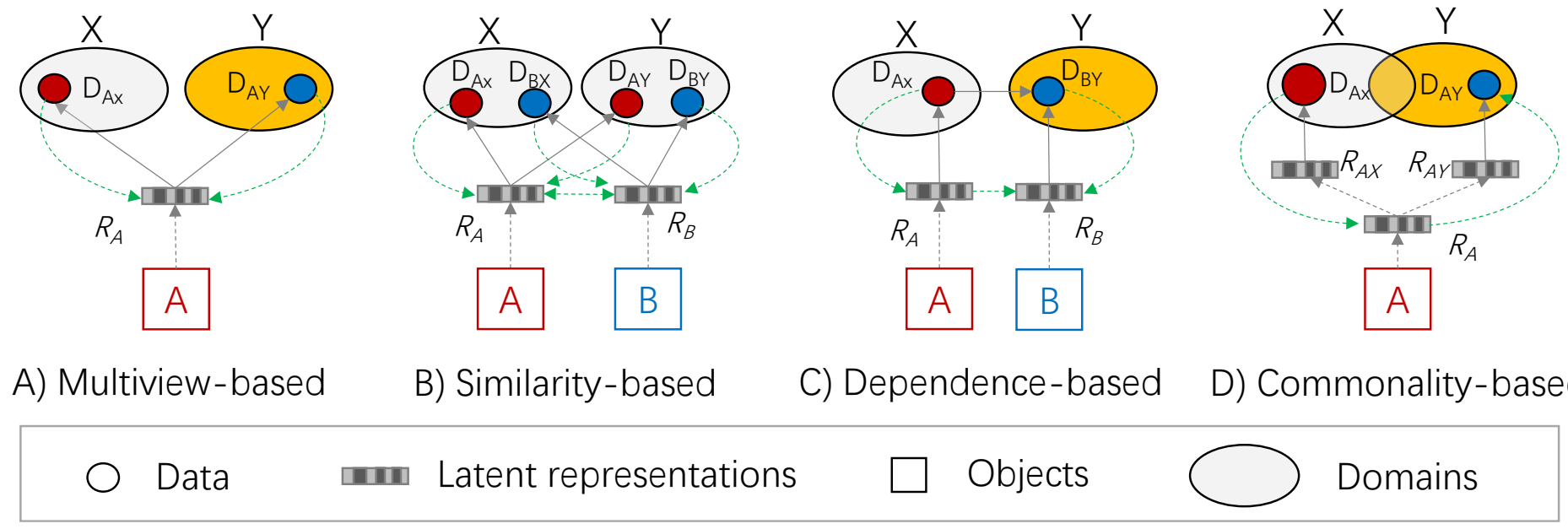

**Figure 8 Four principles of knowledge alignment**

## 5.1 Multiview-Based Principle

### *5.1.1 Rationale of the Principle*

This principle finds different views on the same object in disparate domains to form a better understanding of the object.

As depicted in Figure 8 A), object A contains knowledge denoted by a latent representation $R_A$, which is not explicitly observable. What we can obtain is data $D_{AX}$ and $D_{AY}$ object A generates in domain X and Y respectively. Alternatively, we can say X and Y are two different views observing object A. $D_{AX}$ and $D_{AY}$ can be regarded as records of observation, collectively deriving a better representation of $R_A$ than solely based on each of them. The more disparate X and Y are, the less redundant information about object A the two views offer, and therefore the better representation of $R_A$ is forged. Then, we can solve a problem based on $R_A$.

More specifically, as illustrated in Figure 9 A), $L_X$ and $L_Y$ are latent representations contributed by $D_{AX}$ and $D_{AY}$ respectively. They are a part of $R_A$, collectively forming the knowledge about object A. If the two views are distinct, $L_X$ and $L_Y$ are disjoint. Thus, the knowledge they collectively contribute ($L_{XY} = L_X \cup L_Y$) is maximized, i.e. $L_{XY} = L_X + L_Y$, as shown in the left most case of Figure 9 D). If the two views share something in common, $L_{XY} = L_X + L_Y - overlap$, as depicted in Figure 9 B). The overlap between two views contributes redundant knowledge about object A, which can be derived from either $D_{AX}$ or $D_{AY}$. In the extreme case shown in Figure 9 C), if two views are totally overlapped or one view belongs to the other, the knowledge they contribute to $R_A$ is the least. In other words, the other view is useless, $L_{XY} = \max(L_X, L_Y)$.

In the real world, for example, we can evaluate a student based on their examination results ($D_{AX}$) and sports performances ($D_{AY}$), or even the recommendations from their community neighbors ($D_{AZ}$). To understand a product, we can check its specification, online user evaluations, and sales situations. Those are disparate views observing a student or a product. Thus, the data generated in these domains can be fused to better understand the student and product.

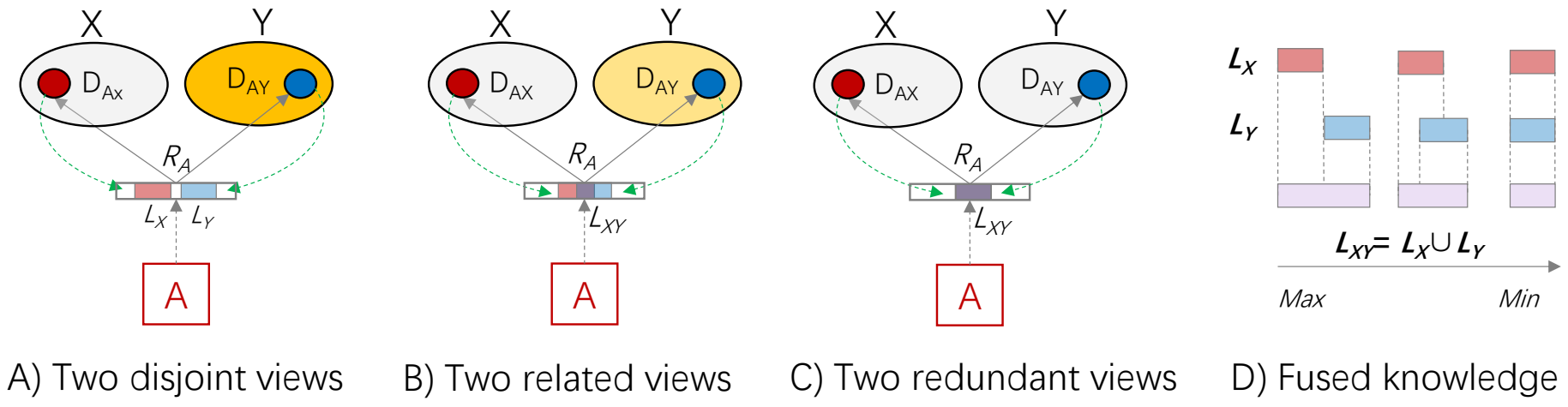

**Figure 9 Performance of knowledge fusion depending on the distinction of the views**

### *5.1.2 Implementations and Examples*

The Multiview-based knowledge alignment has enabled many kinds of models [51], including Co-training [7], Multi-kernel learning [17], subspace learning [10], and deep Multiview learning [53], for a diversity of applications, such as inferring real-time air and water qualities [27][68], forecasting future air quality [70], and predicting flow of crowds in a city [45].

For example, [70] proposes to forecast the air quality of a monitoring station over next 48 hours based on the Multiview learning principle, combining four components consisting of a temporal predictor, a spatial predictor, a dynamic aggregator and an inflection predictor, as depicted in Figure 10. The temporal predictor predicts the air quality of a station in terms of the data about the station, such as local meteorology and AQIs, using a linear regression model. Instead, the spatial predictor predicts a station's future air quality considering its spatial neighbors' data, using a shallow neural network. The results of the two predictors are dynamically combined by the aggregator using a regression tree. Under some unique circumstances, the inflection predictor will be invoked to generate a $\Delta AQI$ which will be appended to the output of the aggregator. In this example, temporal, spatial and inflection predictors are three views on future air quality, implemented by three different machine learning models. The reason of using these models is data scarcity. Given sufficient data, these predictors can also be implemented by deep Multiview models.

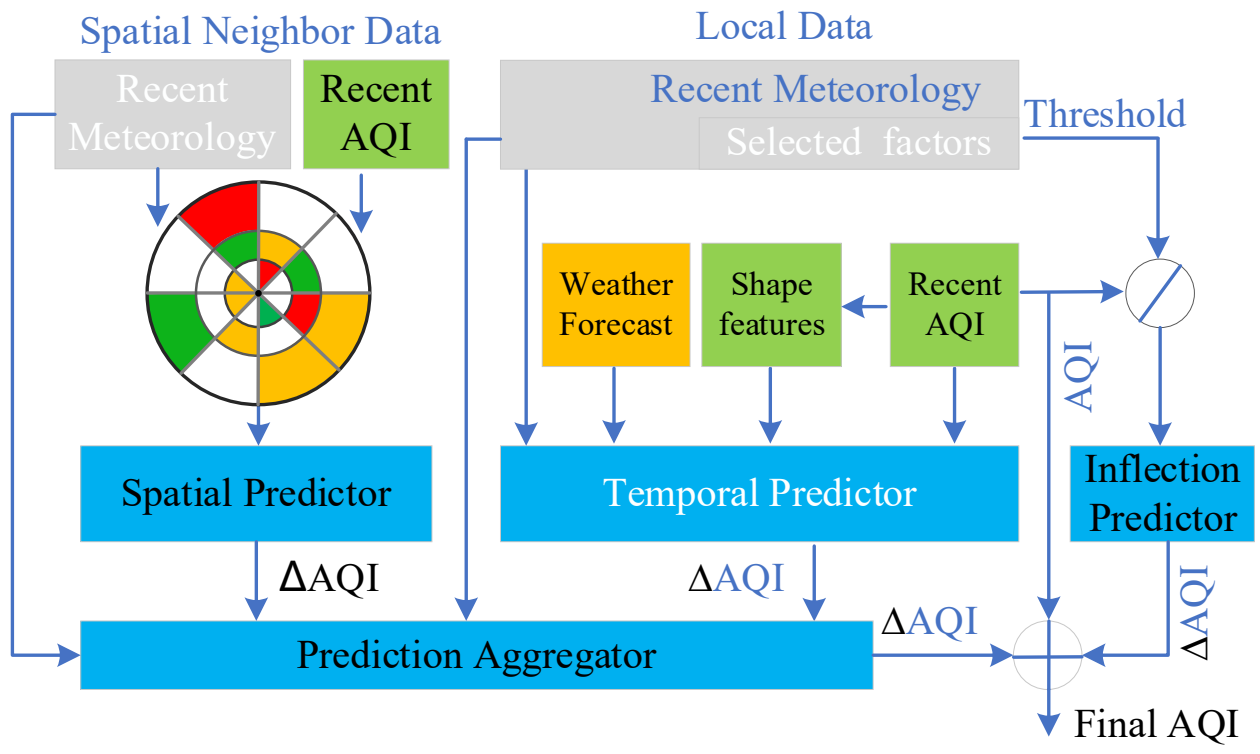

**Figure 10 Forecasting air quality using Multiview-based knowledge alignment principle**

## 5.2 Similarity-Based Principle

### *5.2.1 Rationale of the Principle*

This principle exploits the similarity between objects of the same category and then ensembles their data from different domains to complement the knowledge between each other.

As illustrated in Figure 8 B), let's start with a single domain scenario, where two objects (A and B) have generated data in domain $X$. In many cases, $D_{AX}$ and $D_{BX}$ would be very sparse, thus are hard to form an accurate representation of $R_A$ and $R_B$ respectively. As object A and B belong to the same category, the similarity between them is meaningful and therefore can be used to complement $R_A$ and $R_B$ between each other. More specifically, the knowledge from $D_{AX}$ can complement $R_B$ through the similarity between object A and B. vice versa.

Essentially, this is the original rational of collaborative filtering with which we can infer people's interests in a product they have never seen based on the products they have already bought. The general idea is people who have bought same products in history would have similar interests which would lead to similar shopping behaviors in the future. Thus, we can place different people's purchasing records on different products in a matrix where a row denotes an individual and a column stands for a product. An entry in this matrix represents the times that a specific individual has bought a specific product. By inferring missing entries based on existing records, we can know how likely an individual would purchase an unseen product. This problem can be solved by using a matrix factorization algorithm.

Now, we can extend the scenario to two (or even more) domains. As the data from a single domain is very sparse in many cases, the estimated similarity between two objects is usually inaccurate, therefore reducing the capability of complementing knowledge between each other. Now, in domain $Y$, there are $D_{AY}$ and $D_{BY}$ generated by object A and B respectively. Combining $(D_{AX}, D_{BX})$ with $(D_{AY}, D_{BY})$ will enhance the capability of solving the problem in domain A for two reasons. First, $D_{AY}$ and $D_{BY}$ provide complementary observations for learning a better $R_A$ and $R_B$ respectively. Second, the combination improves the accuracy in estimating the similarity between A and B. The denser $D_{AY}$ and $D_{BY}$ are, the richer knowledge they convey to $R_A$ and $R_B$ is, and the more accurate the estimated similarity is. However, $D_{AX}$ and $D_{AY}$ cannot be simply concatenated, because they have different semantic meanings, representations and resolutions; Neither do $D_{BX}$ and $D_{BY}$. So, they should be placed in different sub-models, such as, matrices or encoders, before aggregated in a sophisticated model.

### *5.2.2 Implementations and Examples*

The similarity-based knowledge alignment principle has motivated a diversity of models, e.g. coupled matrix factorization or tensor decomposition [43][74], and contrastive learning [11].

Coupled matrix factorization and tensor decomposition decompose several matrices and/or tensors (some of them usually are sparse) with shared dimensions together with a constraint of minimizing the recovery errors of non-empty entries, finding a low-dimensional latent representation for each dimension of these matrices or tensors. Then, empty entries of sparse matrices and tensors can be filled with an estimated value based on the productions of those decomposed latent representations. Using these methods, a branch of research works have been done to estimate travel time of a path [50], calculate pollution emission of vehicles on roads [42], diagnose urban noise [69], and conduct location-activity recommendations [74].

For example, Wang et al. [50] employ the similarity-based principle to infer travel time on roads, regarding a road segment as an object $A$. The travel speed of vehicles on a road ($D_{AX}$) and the road's physical properties ($D_{AY}$) are multimodal data from two different domains ($X$ and $Y$). The general idea is similar road segments would have a similar travel speed in the same time interval. However, data denoting travel speeds is very sparse as there are many road segments without being traversed by any sensor-equipped vehicles, e.g. taxicabs with a GPS sensor. That is, the similarity between two road segments $A$ and $B$ solely based on data in domain $X$ is hard to accurate. Thus, $D_{AY}$, including a road's structure and surrounding POIs, from another domain $Y$ is used collectively with $D_{AX}$ to better estimate the similarity between two road segments, and hence improving the inference of travel speed on roads. More specifically, in the implementation, we can deposit $D_{AX}$ in a matrix $M_s$ where a row denotes a road segment and a column stands for a time interval. An entry is the travel speed of vehicles on a particular road segment in a particular time interval. In the meantime, $D_{AY}$ can be placed in another matrix $M_p$, where a row denotes a road segment and a column stands for a kind of physical feature like number of lanes; an entry stores the value of a particular feature pertaining to a road segment. $M_s$ is very spare with many empty entries while $M_p$ is quite dense. By factorizing $M_s$ and $M_p$ together, we can fill empty entries in $M_s$ more accurately than solely based on $D_{AX}$.

Contrastive learning generates latent representations, which maximize their similarity within the same class and minimize it between different classes, for instances in a dataset, by constructing positive and negative examples from the original data. Contrastive learning is a type of self-supervised learning where a model is trained on a task using the data itself to generate supervisory signals rather than relying on externally-provided labels. A series of research [24][28][57] incorporates contrastive learning into graph neural networks to forecast the dynamics of spatio-temporal graphs, e.g. traffic flows on road networks.

## 5.3 Dependence-Based Principle

### *5.3.1 Rationale of the Principle*

This principle utilizes the dependency between different objects' properties to reinforce the knowledge between each other.

As illustrated in Figure 8 C), A and B are two objects of different category without a meaningful similarity between them. However, the data, $D_{AX}$ and $D_{BY}$, they generate in domain X and Y respectively may have a probabilistic dependency, revealing the interactions between $R_A$ and $R_B$. The dependency provides contexts and constraints for a more accurate estimation of $R_A$ and $R_B$, which leads to a better performance of task completion. The stronger dependency between two objects, the richer knowledge one can complement to the other.

### *5.3.2 Implementations and Examples*

This principle has been widely used to design structures of probabilistic graphical models, like Conditional Random Field (CRF) and Latent Dirichlet Allocation (LDA) [6], and deep neural networks including self-attention mechanisms [48][56][61], for travel speed estimation on roads [42], crowd flow prediction in urban regions [19][61], air quality forecast [56], functional zone inference in a city [58], and geo-sensory data prediction [25]. For example, vehicles' travel speed on a road depends on the road's properties, such as the number of lanes, speed limits, indegrees and outdegrees. The speed also depends on that of adjacent road segments and weather conditions. By placing these factors into a probabilistic graphical model, Shang et al. [42] infer the traffic volume on each and every road segment in a city given a few road segments with observations of travel speed.

As shown in Figure 11 A), Yi et al. [56] designed a deep neural network-based model to forecast air quality, after sufficient air quality data has been accumulated. Weather forecast, current meteorology, concentrations of other air pollutants, and time of day are regarded as implicit factors impacting air quality. Thus, they are thrown into separate fusion nets with AQIs to generate individual predictions which are then merged by a hidden layer of neural network to derive a final result.

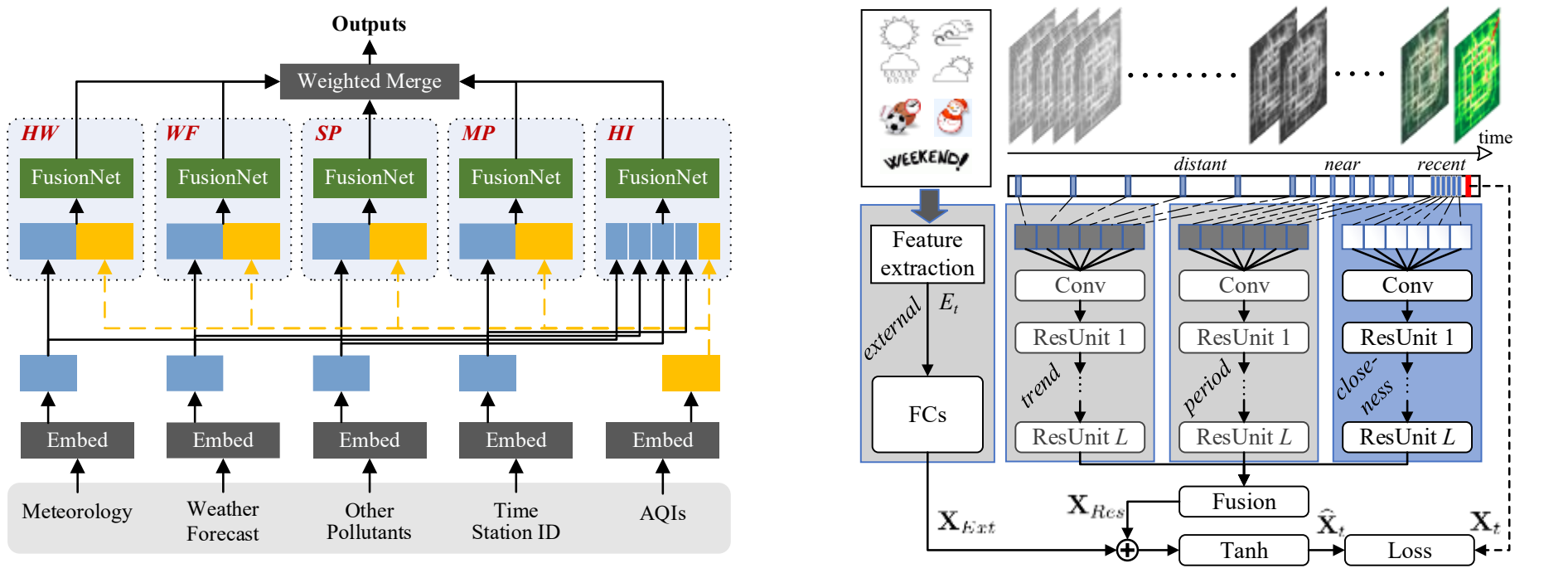

A) Forecast air quality using deep neural networks

B) Predict crowd flows using spatio-temporal residual networks

**Figure 11 Examples of models using dependency-based knowledge alignment principle**

As illustrated in Figure 11 B), using the dependency-based knowledge alignment principle, Zhang et al. [61] propose the first deep learning model dedicatedly designed for spatio-temporal data to predict the flow of crowds in a city. The in-and-out crowd flows in a region depend on the flow in the region over recent times, near histories and distant histories, because crowd flows as a kind of spatio-temporal data have temporal closeness, periods and trends properties. In addition, they also depend on flows of their spatial neighbors as well as weather conditions and major events. Considering these factors, the proposed spatio-temporal residual networks only select the data from a couple of time frames, such as recent hours, the same time of yesterday and that of last week, to predict the flow at future times, whereas other frames are neglected to reduce the complexity of model structure. The data of recent times, near histories and distant histories are fed into three sub-networks to model the temporal closeness, periods and trends properties respectively. In each sub-network, a residual convolutional neural network is constructed to capture the spatial correlation between flows of regions with different distances in between. The impact of weather conditions and major events are finally appended to the aggregated fusion results of the three sub-networks through a full connect network.

## 5.4 Commonality-Based Principle

### *5.4.1 Rationale of the Principle*

This principle exploits the commonality shared by different domains, leveraging the data generated by an object in one domain to enrich the knowledge of data the object creates in other domains.

As illustrated in Figure 8 D), object A's latent representation $R_A$ generates two finer-grained representations $R_{AX}$ and $R_{AY}$ which denote A's knowledge in domain $X$ and $Y$ respectively. $D_{AX}$ and $D_{AY}$ are observations of $R_{AX}$ and $R_{AY}$. As domain $X$ and $Y$ share something in common, $R_{AX}$ and $R_{AY}$ would share some common knowledge derived from $R_A$. Particularly, when $D_{AX}$ is abundant while $D_{AY}$ is sparse, we can leverage $R_{AX}$ learned from $D_{AX}$ to consolidate $R_A$ and thus enhance the capability of generating $R_{AY}$, which would further improve the accuracy of generating $D_{AY}$. That is, their knowledge can complement each other through collectively constructing a better $R_A$. The more commonality two domains share is, the richer knowledge can be transferred between the two domains.

More specifically, as depicted in Figure 12 A), $L_{XY} = L_X \cap L_Y$ denotes the commonality shared by domain X and Y. $L_X$ is a part of $R_{AX}$, derived from $L_{XY}$. Likewise, $L_Y$ is a part of $R_{AY}$, derived from $L_{XY}$ too. The more domain $X$ and $Y$ share, the richer knowledge $L_{XY}$ contains, which is represented by a larger size of purple block. On the contrary, the less domain $X$ and $Y$ share, the smaller $L_{XY}$ is, as depicted in Figure 12 C). As $L_{XY}$ is a bridge connecting $L_X$ and $L_Y$, a larger $L_{XY}$ can provide a wider bandwidth for transferring knowledge from $D_{AX}$ to $D_{AY}$.

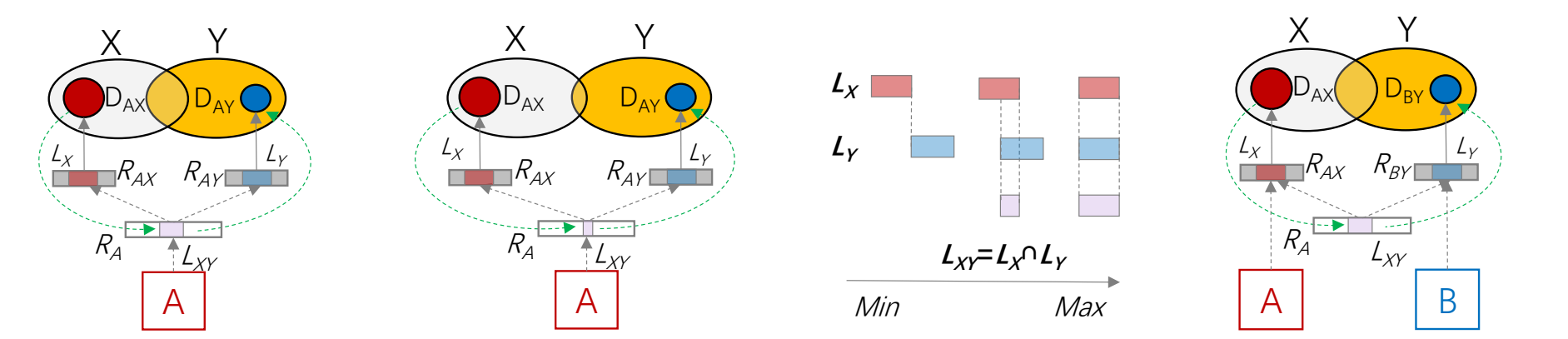

**Figure 12 Performance of knowledge fusion depending on commonalities between domains**

For example, if we want to provide travel recommendations to users, we need to collect enough data $D_{AY}$ from travel agents or websites. However, $D_{AY}$ is relatively hard to obtain as many people do not have a record in these data sources. Fortunately, we may have people's records $D_{AX}$ in purchasing or browsing books in online websites. There are some commonalities between an individual's interests in books and tourist attractions. Hence, we can leverage $D_{AX}$ to improve the prediction on $D_{AY}$.

The commonality-based principle can be further extended to two and more objects of different categories. For example, as illustrated in Figure 12 D), object $A$ and $B$ are two different categories of animals whose similarity is not meaningful. However, the two tasks classifying them into a category have a certain commonality, e.g. people would recognize an animal by checking some key features like their heads, ears, eyes, furs, arms and legs. $R_{AX}$ can be regarded as meta knowledge of the task abstracted from $D_{AX}$. Though the features of $A$ and $B$ are different, the common knowledge $L_{XY}$ shared by the two tasks can convey from one (with rich data) to another (with a few data), i.e. from $R_{AX}$ to $R_{BY}$. Then, the task classifying $B$ is enhanced with external prior knowledge from $R_{AX}$.

*5.4.2 Implementations and Examples*

Based on the commonality-based knowledge alignment principle, a series of research themes, such as multitask learning (MLT) [4][9], transfer learning [35], and meta learning [16][20], have been carried out in the past decade. Sometimes, MLT is regarded as a kind of transfer learning [35]. Some researchers also take transfer learning as a special case of meta-learning [23].

Multitask learning learns a problem together with other related problems at the same time, using a shared representation. This often leads to a better model for the main task, because it allows the learner to use the commonality among the tasks. MTL works well if these tasks have some commonality and are slightly under sampled. For example, Liu et al. [27] forecast the water quality of multiple stations simultaneously using a multitask learning framework, obtaining a 15% improvement on predictive accuracy over a stand-alone prediction on individual stations. More examples can be found in our previous review on cross-domain data fusion [65].

Transfer learning consists of two categories: transductive and inductive learning according to whether label data are available in source and target domains. Transductive learning handles cases where the task is the same but the source and target domain are different. For instance, Yang et al. [54] propose heterogenous transfer learning to improve image classification using text classification, based on the semantic commonality between images and its surrounding texts. Inductive learning handles cases where tasks are different in source and target domains. For example, Zhang et al. [62] simultaneously predict in-flow in a region and the transition between regions using a multitask learning-based deep neural networks.

Meta learning aims to improve adaptation through transferring generic and accumulated knowledge (meta-data) from prior experiences with few data points to adapt to new tasks quickly without requiring training from scratch [16][20]. Meta-learning is most commonly known as "learning to learn". The example shown in Figure 12 D) can also be regarded as a representative of meta learning. Using meta learning algorithms, a diversity of research has been done for predicting purchasing orders during shopping festivals [37], traffic flow in fine-grained locations throughout a city [36], and service time of delivery tasks [41].

### 5.5 Differences and Connections between these Principles

**1) Differences between Multiview-based and commonality-based Principles**: Though Figure 9 and 12 look similar, the differences between them are two folds. First, the Multiview-based principle aggregates $L_X$ and $L_Y$ to collectively form a better $R_A$, i.e., $L_{XY} = L_X \cup L_Y$, while the latter principle leverages the commonality between two domains to transfer knowledge from one domain to another, i.e. $L_{XY} = L_X \cap L_Y$. Thus, the more correlated two domains are, the less useful the Multiview-based principle is while the more effective the commonality-based one is. Second, the former principle solves a problem based on $R_A$ whereas the latter uses $R_{AY}$ or $R_{BY}$ to complete a task. $L_{XY}$ is a bridge delivering complementary knowledge from $R_{AX}$ to $R_{AY}$.

**2) Differences between similarity-based and commonality-based Principles**: The differences consist of two parts. First, the former principle utilizes the similarity between two objects of the same category, while the latter employs the commonality between the observations of an object in two (related but different) domains or that between two (related but different) tasks involving multiple objects of different categories. Second, the former principle improves the latent representation of the target object ($R_A$), whereas the latter enhances the latent representation of an object in the target domain ($R_{AY}$) or other object' representation in a target task ($R_{BY}$).

**3) Connections between these principles**: the four knowledge alignment principles can be employed together in an AI task.

- Combing the commonality-based with Multiview-based principles: Liu et al. [27] infer urban water quality by combining multitask learning with a Multiview model. Pan et al.[36] integrate meta learning with Multiview learning to predict urban traffic.

- Combing the dependency-based with Multiview-based principles: Liang et al. [25] incorporate attention mechanisms in a Multiview learning model to predict geo-sensory data. Zhang et al. [62] predict flow of crowds in a region, using major events and weather conditions as impacting factors in a Multiview-based deep learning model.
- Combing the similarity-based with Multiview-based principles: Liu et al. [28] consider spatial and temporal views in a contrastive learning framework to predict the dynamics of spatio-temporal graphs. Similar idea is also employed in [38] to forecast urban flows.
- Combining the dependency-based with commonality-based principles: Ruan et al. [41] employ a transformer-based representation layer (encoding delivery circumstances) in a meta learning framework to predict the service time of delivery tasks.

## 6. KNOWLEDGE FUSION PARADIGMS

After determining the links between multimodal data, we can start designing specific AI models based on two knowledge fusion paradigms, which is comprised of a precise fusion and a coarse fusion.

### 6.1 Definitions and Differences between the two Paradigms

**The precision knowledge fusion paradigm** first extracts *precise* knowledge from each piece of data as accurately as possible through precise data transformation approaches, such as map-matching [29], image segmentations and entity extractions. These precise knowledges are then *explicitly* connected based on the previously determined links between data in some interpretable AI models like knowledge graphs, collaborative filtering, and probabilistic graphic models.

**The coarse knowledge fusion paradigm** first generates intermediate representations of multimodal data, which can be regarded as *coarse* and preliminary knowledge, using some coarse data transformation methods, such as text embedding and image encoders. The coarse representations of knowledge are then *implicitly* connected, most likely in a deep learning model, based on the links between data.

**Differences**: The differences between the two knowledge fusion paradigms are two folds. One is the *precise* knowledge versus *coarse* knowledge in the first step. The other is the *explicit* connections versus *implicit* connections in the second step.

Note that the two knowledge paradigms are not specific models, though we use a coupled matrix factorization and a deep neural network for demonstrating the idea in the following sections. Both paradigms are useful and should be employed properly based on application scenarios. It is vital to select a right knowledge fusion paradigm before designing specific model structures.

The precise fusion paradigm should be employed when the following three requirements are satisfied: 1) we have a relatively clear understanding of the problem; 2) the data is insufficient; and 3) the extraction of preliminary knowledge is accurate enough. On the contrary, if the understanding of a problem is vague, the data is rich, and the extraction of the preliminary knowledge is hard to be accurate, the coarse knowledge fusion paradigm should be considered.

**Connections**: The two knowledge fusion paradigms can be combined to complete complex tasks. For example, combing knowledge graphs and large language models to answer complicated questions [14], like “how many people with an age from 20 to 30 have visited art museums in downtown areas”, or “how many senior people live alone in community A”. Particularly, with advanced methods that can automatically build accurate and large-scale knowledge graph [71], we are more likely to harness the two fusion paradigms in solving one problem.

**Examples:** Figure 13 presents an example of the two knowledge fusion paradigms. There are three categories of datasets which have different data modalities and look disparate. Once we know the three categories of datasets were generated by the same person in different data sources, we can find anchor points for aligning the knowledge from these datasets. More specifically, the three datasets can be regarded as different observations on a fact, i.e. a person $P$ lives in a location $L$ (denoted as $P\ lives\ in\ L$), through three different views. For example, because living in location $L$, person $P$ fills her home address in an online form, generating structured data in E-government services, depicted as the left most dataset in Figure 13. Because $P$ lives in $L$, her vehicle generates a few GPS trajectories starting from and ending at the location, shown as the middle dataset. For the same reason, $P$ would share photos and tweets about the environment and buildings around $L$, generating the rightmost dataset. In other words, the three datasets are different presentations of the same knowledge ($P\ lives\ in\ L$), and thus can be fused to achieve the same goal, e.g. to better understand $P$ and $L$ as well the link between them.

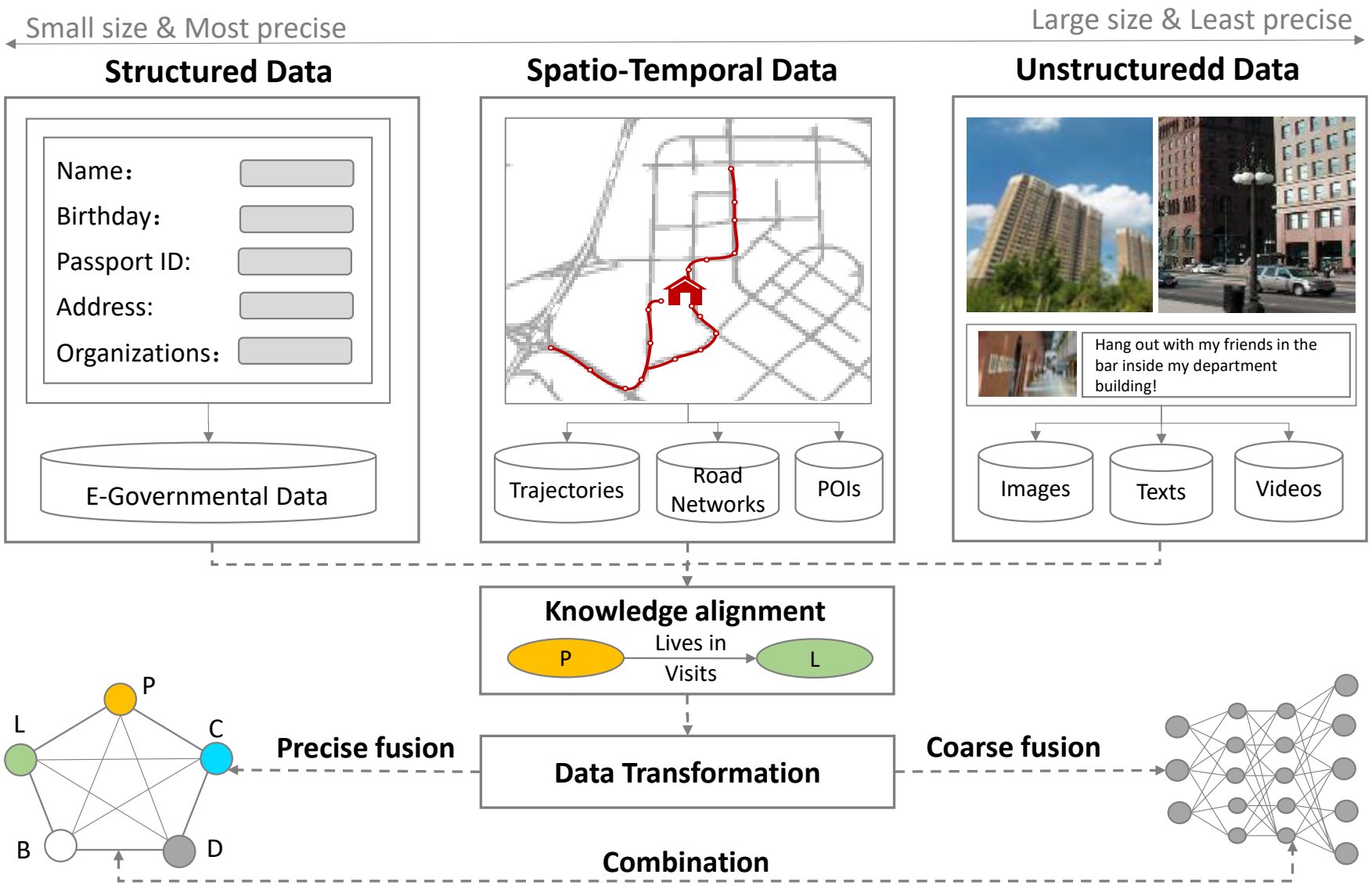

**Figure 13 Example of two knowledge fusion paradigms**

In reality, there are a lot of people generating these three categories of datasets in many locations (not limited to their homes, e.g. $P\ visits\ L$) in the three data sources. However, they may have insufficient data in each data source. For instance, some people may not fill in accurate addresses, or did not record many trajectories, or post few tweets about these locations. At this moment, we need to fuse these three datasets if wanting to have a better prediction on the link between an individual and a location. The strength of a link could have different semantic meanings in different applications, e.g. how likely an individual would be interested in a location, or the probability that an individual could have visited a location, or the likelihood of a person working at or living in a location. Those are ordinary tasks in location recommendations, human behavior recognitions, and user profiling.

## 6.2 Using the Precise Knowledge Fusion Paradigm

Regarding the example presented in Figure 13, we first extract precise knowledge about people and locations from the three datasets through precise data transformation approaches, which will be detailed in Section 7. For example, we can build links between an individual and locations by matching the address they have filled in the first data source against addresses in a POI database. If matched, a link is built between the individual and a specific POI. In the meantime, we usually detect stay points [73] from an individual's trajectories and then match these stay points against a POI database based on their GPS coordinates. Typically, the closest POI to a stay point is selected as the exact location where the individual has stayed. Then, a link is created between the individual and the POI. In addition, we need to extract names of POI entities from the individual's tweets or recognize specific POIs based on photos posted by the individual.

Then, we can construct an explicit knowledge graph between people and locations, where a link could denote "*lives in*", "*visits*", or "*works at*" etc., as shown in the bottom left part of Figure 13. Likewise, the links between individuals and organizations can be explicitly built and then added into the knowledge graph, where a link could represent "*works for*" or "*managing*" etc. After that, we can use classification algorithms proposed for heterogenous information networks [46] to label the category of each link or even discover underlying links between different nodes using link prediction algorithms.

For instance, as shown in Figure 14, we can explicitly represent the links between users and locations in matrix $M_x$ where each entry $e_{pl}$ denotes the strength of the link between user $p$ and location $l$. Values of $e_{pl}$ can be precisely obtained from the three categories of datasets through the approaches mentioned above. $M_x$ is intrinsically sparse because people would have not visited many locations. Additionally, these datasets are just a small sample of people's real life.

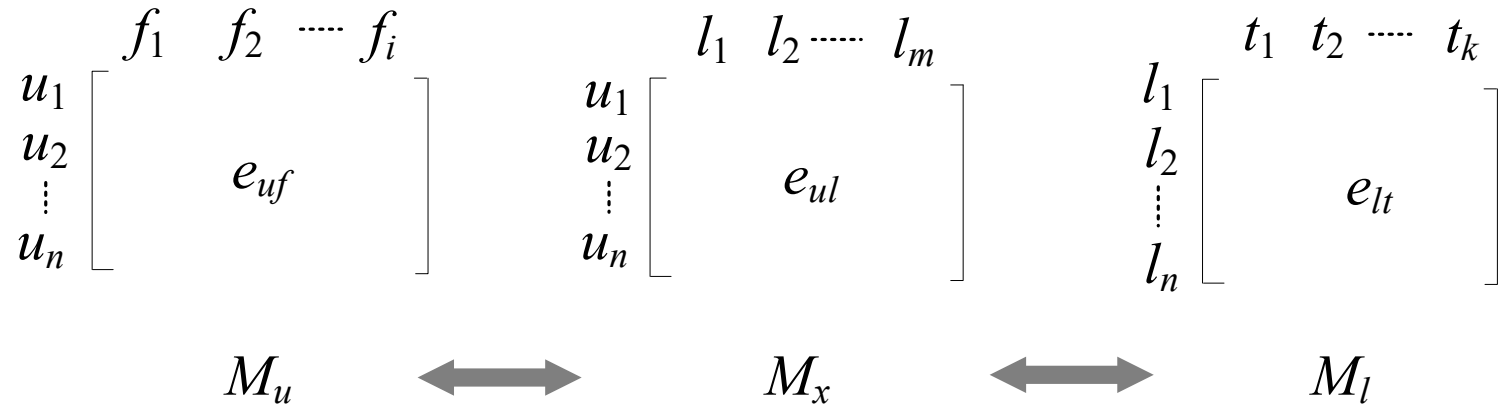

**Figure 14 Example of precise knowledge fusion paradigm**

As a result, we can deposit each user's profile from the structured dataset into matrix $M_u$, where each entry $e_{uf}$ denotes user $u$'s feature in profile field $f$. In the meantime, we can build matrix $M_l$ from the unstructured dataset, where entry $e_{lt}$ stands for the presence of tag $t$ generated by users on location $l$. $M_u$ and $M_x$ share the user dimension, and $M_l$ and $M_x$ share the location dimension. As matrices $M_u$ and $M_l$ are much denser than $M_x$, the knowledge they contained about users and locations can be transferred to $M_x$ for inferring its missing entries by using a coupled matrix factorization.

### 6.3 Using the Coarse Knowledge Fusion Paradigm

As we can see in Figure 13, the leftmost dataset, i.e. the structured data, has the least size for each piece of data while containing the most precise knowledge. The rightmost dataset, i.e. the unstructured data, has the largest size for each piece of data whereas containing the least precise knowledge. This calls for sophisticated algorithms for extractions and recognitions, which are indeed hard to be precise. Thus, the coarse knowledge fusion paradigm is proposed, first turning multimodal data into intermediate representations using different kinds of encoders or embedding algorithms.

Following the example shown in Figure 13, Figure 15 presents an example of the coarse knowledge fusion paradigm, where four different encoders (depicted as four different colors) are designed to generate intermediate representations for an individual's images, texts, trajectories and spreadsheet data are generated respectively. The intermediate representations of the four datasets are then aggregated through some hidden layers, generating a latent representation about the individual's behaviors and interests. After processed by another set of hidden layers, the latent representation will be turned into a vector of output where each entry could denote the linking strength between the individual and a specific location. Sometimes, the ideal of Mixture-of-Experts (MoE) [34] can be employed in this paradigm, activating different components (or experts) in a large model according to inputs to obtain a more accurate result with less resources.

Not that what Figure 15 presents is just an example of the coarse knowledge fusion paradigm. If a model first turns multimodal data into latent representations and then fuse them in a deep neural network-like model, this model falls in the coarse knowledge fusion paradigm. Models discussed in [15][44][63][64] can be regarded as instances of this paradigm. Once professionals have gone through the Domain and Link layers shown in Figure 5, i.e. knowing "what to fuse" and "why can be fused", they can select a knowledge fusion paradigm and then design specific model structures, which might be very different from the one shown in Figure 15, for solving their own problems. The specific model structures are not the focus of this paper, as there are many models proposed in literature.

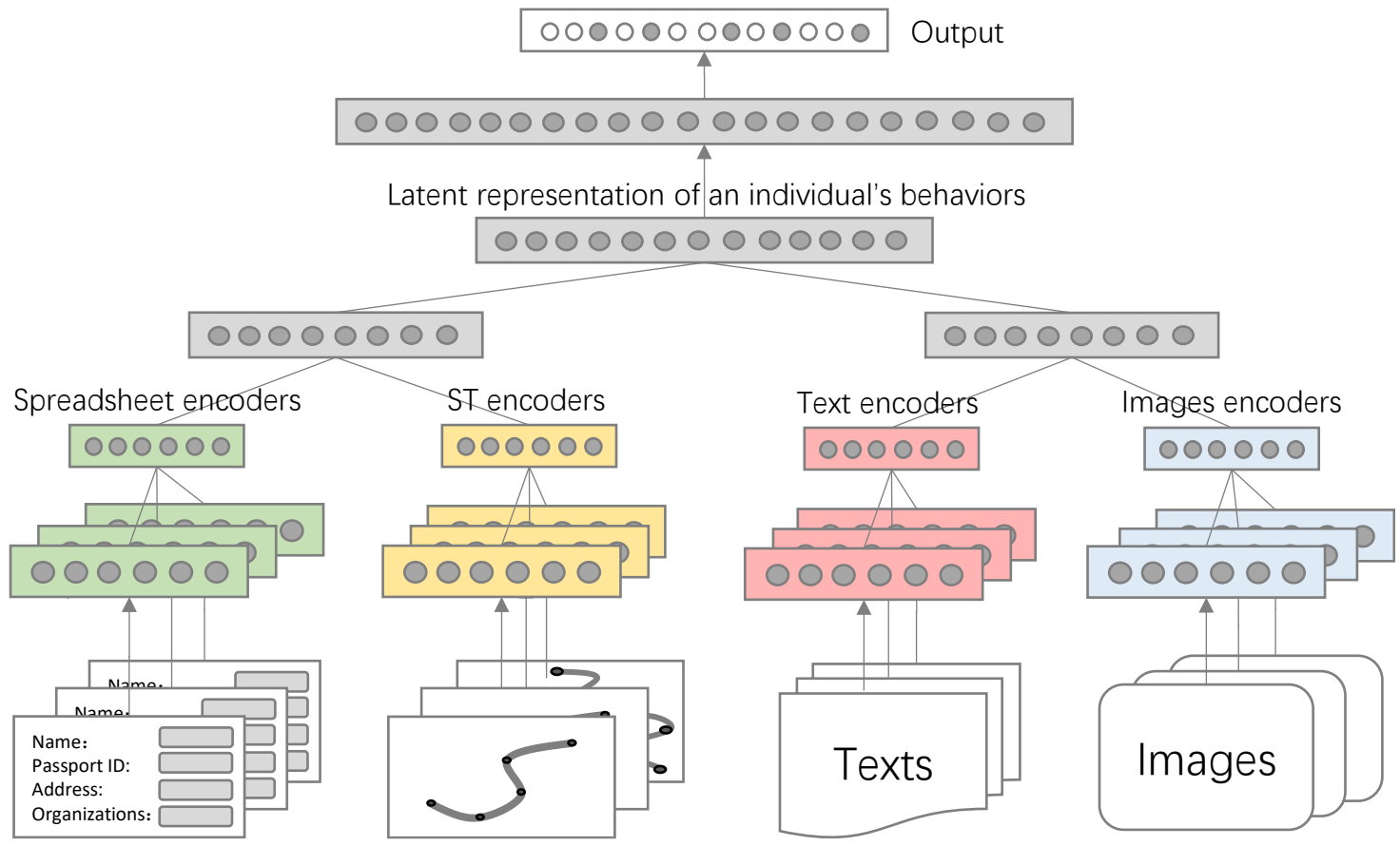

**Figure 15 Example of coarse knowledge fusion**

## 7. DATA TRANSFORMATION

### 7.1 Motivations of Data Transformation

Data of different modalities has different structures, scales, resolutions and distributions, preventing them from being processed directly by AI models.

1) Data with different structures require different encoding algorithms to turn them in to a consistent representation before sending them to an AI model. For example, as illustrated in Figure 16 A), road networks and POIs have different data structures, represented by a spatial graph and spatial points respectively. Without a data transformation process, no AI models can take them as an input directly and simultaneously, let alone fusing them with texts and images. In addition, the original representation to align multimodal data is usually very sparse, e.g. using a matrix to represent different networks or using a one-hop vector to denote different terms, resulting in a high order space for a data instance. This does not only increase computational loads but also cause troubles to training processes.

2) Data with significantly different scales would result in a vulnerable training process. For instance, as depicted in Figure 16 B), features extracted from two trajectories with tremendously different lengths will have extremely large or small values. Some instances with extreme values may be treated as outliers and thus be ignored by some AI models. Alternatively, if there are many extreme values in the input, a training process may be faced with exploding gradient or vanishing gradient problems if an artificial neural network-based model is selected.

3) Data with different resolutions requires different structures of AI models, such as the size of input and numbers of hidden layers. As shown in Figure 16 C), for example, two moving objects traversing the same route will generate two different trajectories if using different spatial and temporal sampling rates, e.g. recording a GPS point every 5 seconds and 10 minutes respectively. That is, they have different spatio-temporal resolutions (a.k.a. granularities), containing different numbers of records, which call for different numbers of units in an LSTM. Additionally, even if the two trajectories have the same resolution, the GPS points they contain may not be comparable either for two reasons. One is GPS coordinates are recorded as double variables, which cannot go through an equal judgement operation. Second, the two moving objects passing the same location would generate different GPS coordinates because of positioning errors or being positioned by different satellites. Thus, it is hard to through these two trajectories into the same AI model.

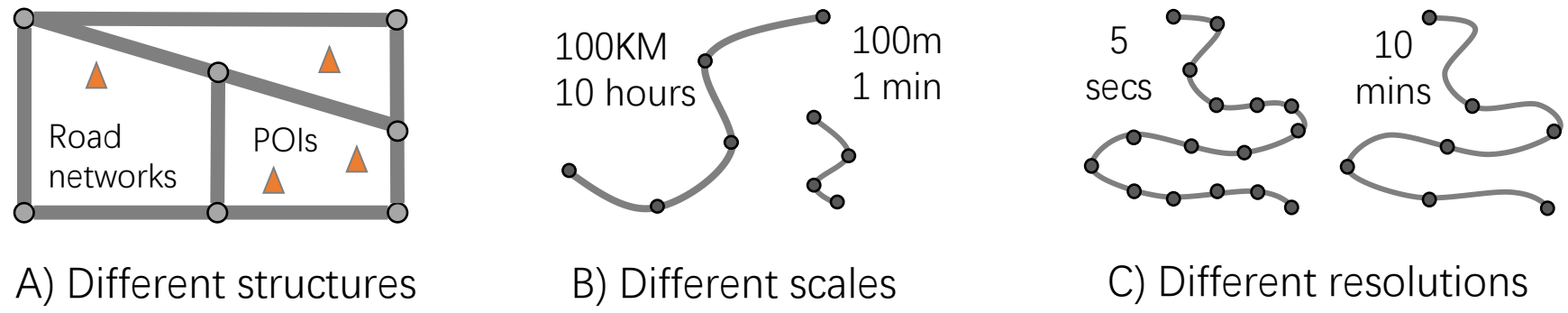

**Figure 16 Differences between multimodal data**

4) Data with different distributions on class labels or spatial and temporal spaces would mislead the inference of an AI model. For example, an image corpus has significantly more photos about a class label than others. Likewise, a trajectory corpus may have GPS traces generated by different transportation modes, such as cycling, driving and walking. The number of instances varies in different modes. These imbalanced distributions will lead to redundant training data for some classes, dominating the training process and compromising the overall effectiveness of the model.

As a result, before sending selected multimodal data into a designed AI model, we need to transform them into a consistent representation, tackling the challenges caused by different structures, scales, resolutions and distributions.

### 7.2 Key Factors for Designing Data Transformation

The data transformation process depends on two main factors.

One is the intrinsic properties of data modality, which denotes the nature of datasets to be processed regardless of applications. For example, as shown in Figure 17, spatio-temporal data has unique properties consisting of spatial distances, spatial hierarchies, temporal closeness, temporal period patterns and temporal trends. When used correctly, these properties cannot only reduce the complexity of model structures but also improve the accuracy of an AI model. Thus, the data transformation algorithms should preserve these properties in the generated representations. In other words, different data modalities should have different representations generated by different transforming algorithms.

The other is the links between multimodal data, depending on problems, domains and the philosophy of knowledge alignment (as shown in procedure ⑥). This can be regarded as data's adaptions to applications. That is, data of the same modality could have different representations generated by different transforming algorithms in different categories of application scenarios.

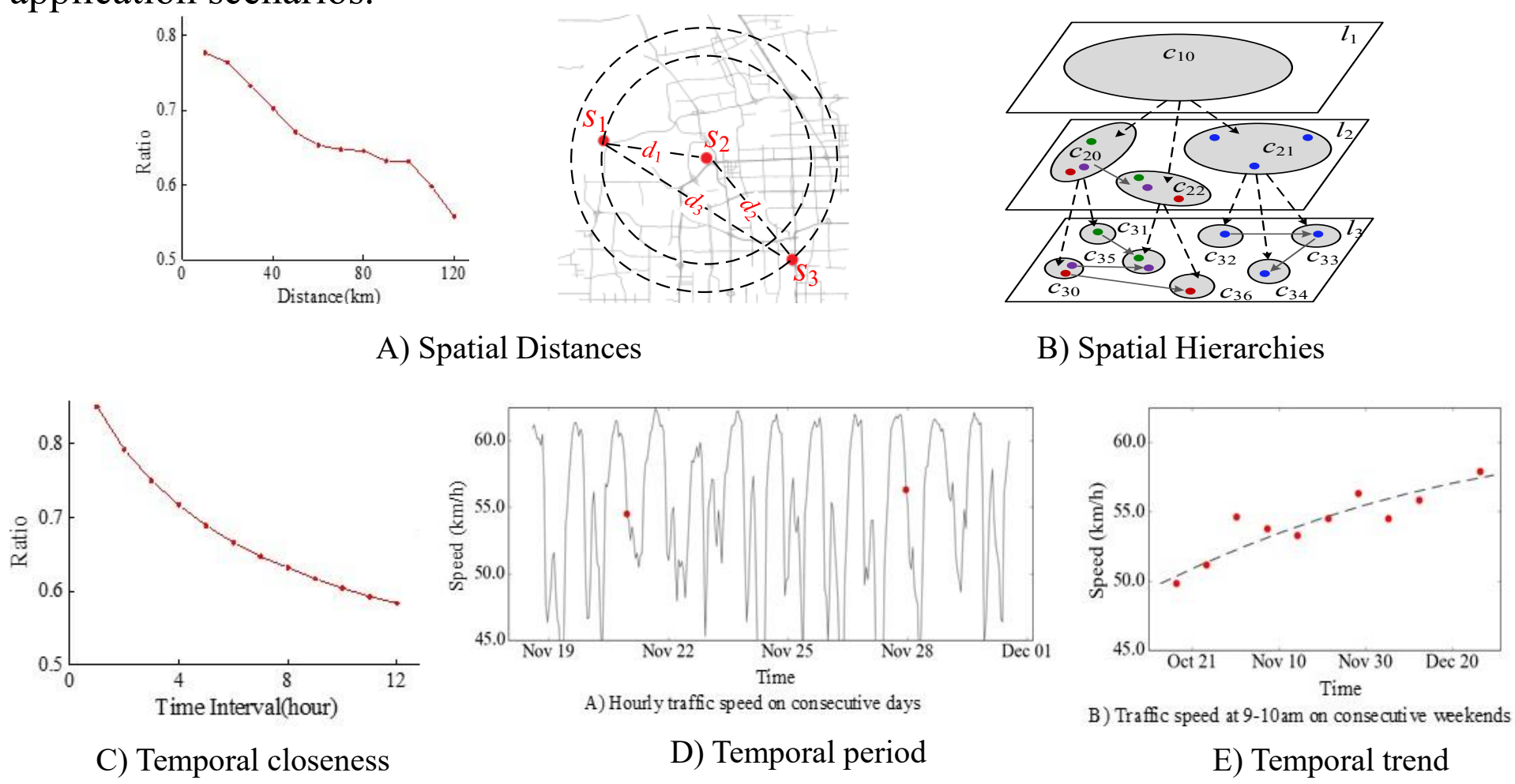

**Figure 17 Unique properties of spatio-temporal data**

### 7.3 Architecture of Data Transformation

As shown in Figure 18, the architecture of data transformation is comprised of three components: data preprocessing, precise transformation and coarse transformation, generating three data transformation approaches (①, ② and ③). The data preprocessing is a foundation, tackling the challenges posed by different data distributions, scales and resolutions. The latter two transformation components mainly solve the problem caused by different data structures and data sparsity. Combined with the data preprocessing, two transformation components formulate a precise transformation approach (denoted as ①) and a coarse transformation approach (denoted as ②) respectively. Sometimes, a coarse transformation process can be performed after a precise transformation, generating a hybrid approach illustrated as ③. The selection of data transformation approaches depends on the features of applications and nature of data modalities.

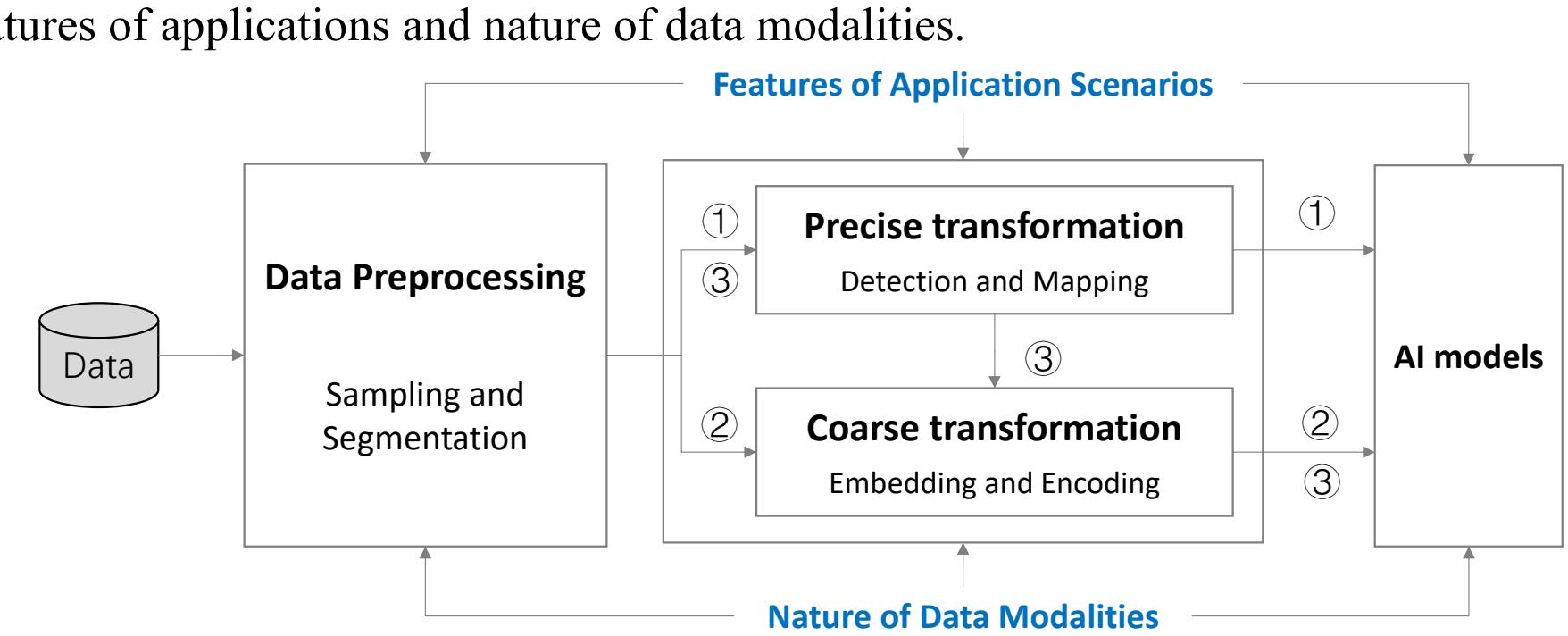

**Figure 18 Framework of data transformation**

### *7.3.1 Data Preprocessing*

This component consists of data sampling and segmentation methods.

1) **Data sampling** methods select a necessary number of instances from an original dataset for a uniform distribution of class labels, tackling the challenges posed by different data distributions. It scales down the size of a dataset by skipping redundant data, and thus reduces computational workloads. These sampling methods also convert instances of different resolutions into comparable ones for a consistent computing process, solving the problem caused by different data resolutions. For example, sampling photos of different resolutions down to a consistent pixel level, or sampling trajectories of different recording frequencies up or down to the same temporal granularity.

**2) Segmentation methods** partition data into different fragments so that each fragment has a same level of scale or fragments of interest can be identified. As shown in Figure 19 A), when processing documents with a very long length, we typically partition them into some portions of a similar length. A trajectory can be divided into uniform segments by spatial distances $d$ or temporal span $t$, as illustrated in Figure 19 B). After that, trajectories with different spatial and temporal scales can be processed together. As shown in Figure 19 C), to better analyze important objects, e.g. people or building, we need to segment areas of interest from the background of an image using image segmentation methods.

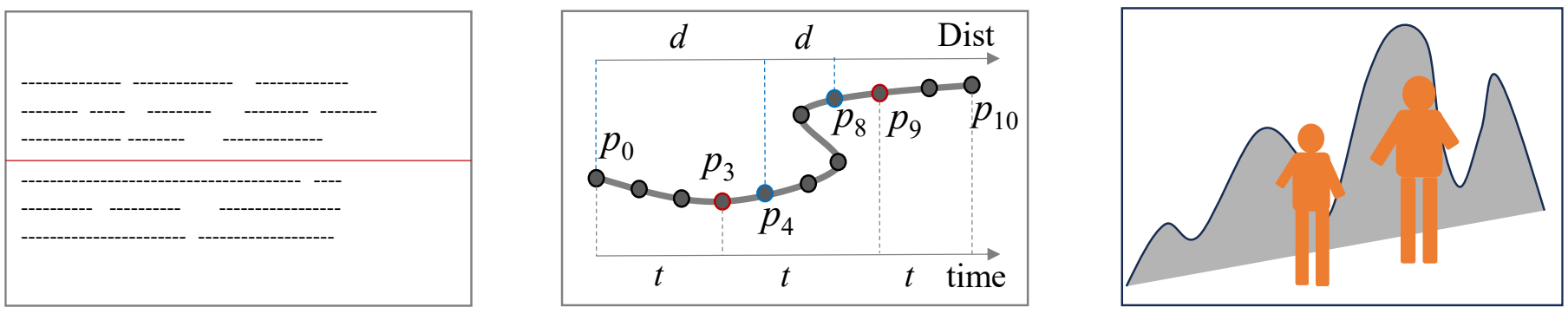

A) Text segmentations B) Trajectory segmentations C) Image segmentations

**Figure 19 Examples of data segmentation**

### *7.3.2 Precise Data Transformation*

This component is comprised of a detection and mapping methods varying in data modalities.

1) The detection methods extract key elements from multimodal data and then uses these elements to represent the original data.

For example, as illustrated in Figure 20 A)-①, we can detect stay points, like $s_1$ and $s_2$, where a user has stayed for a while, from a given trajectory $Tra$. Then, this trajectory can be represented by a sequence of stay points, e.g. $s_1 \rightarrow s_2$, which carry more semantic meanings, such as shopping, having a dinner, or hanging out at some places, than original GPS points. Further, using spatial clustering algorithms, we can detect clusters of stay points, e.g. $C_1$ and $C_2$ shown in Figure 20 A)-②, from multiple trajectories. Then, trajectory $Tra$ can be represented by $C_1 \rightarrow C_2$. Additionally, as illustrated in Figure 20 A)-③, a hierarchy of clusters can be detected from many people's stay points by using hierarchical clustering algorithms. A corpus of stay points is iteratively divided into sub-clusters, forming a spatial hierarchy where the scope of a parent cluster is the aggregation of its children's spatial areas.

Likewise, we can detect entity names from a given document based on a dictionary, as depicted in Figure 20 C), using entity extraction algorithms. Then, this document can be represented by a bag of words with different term frequencies (TF) and inverse document frequencies (IDF). Similarly, as shown in Figure 20 D), we can detect people's faces from a given image and then match these faces against a facial database for entity recognition.

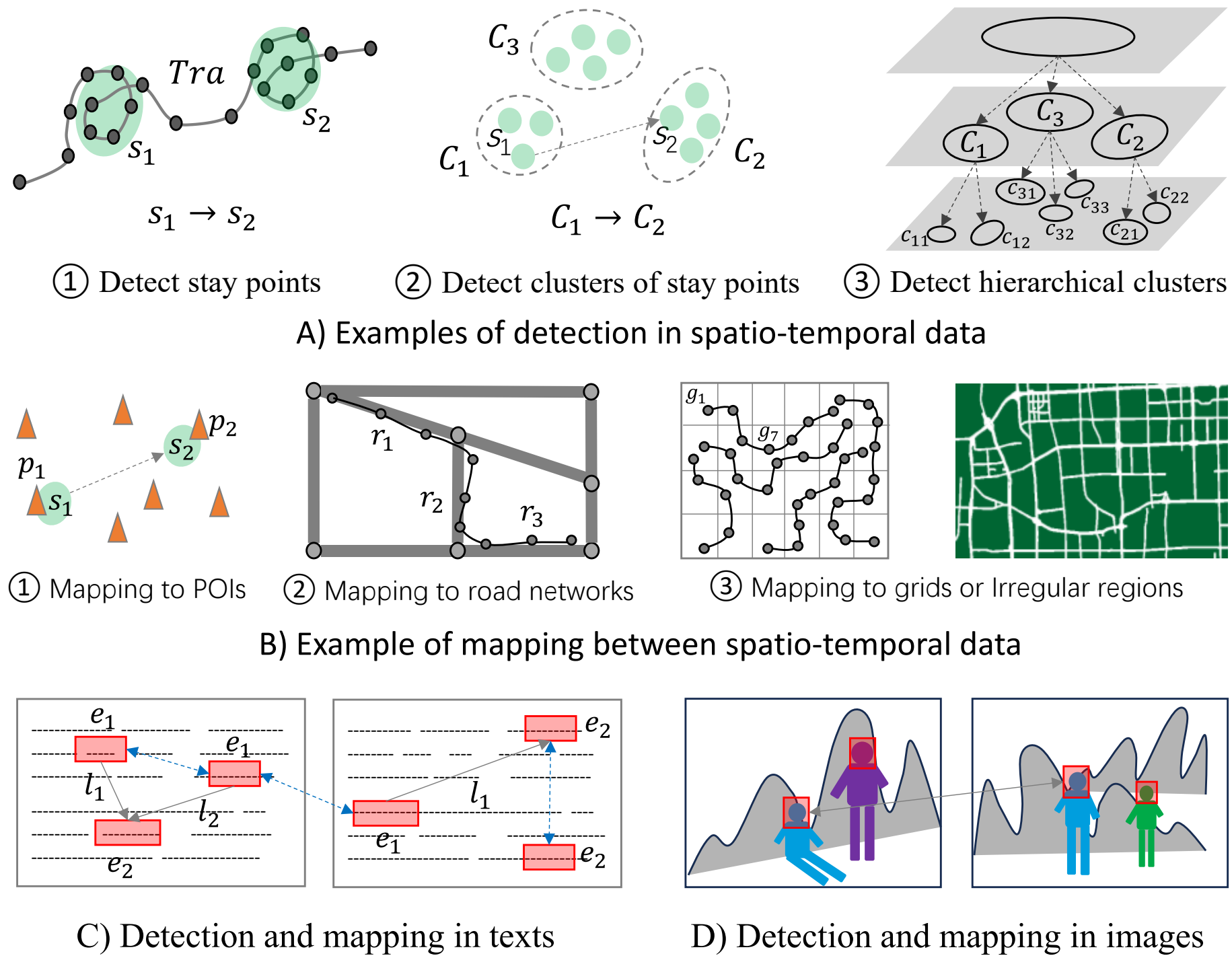

**Figure 20 Examples of detection and mapping**

2) The mapping methods project data onto a shared framework or matching them against a common foundation, generating a simplified and consistent data representation. For example, as illustrated in Figure 20 B)-①, to identify the exact location that a user has been to, we can map their stay points $s_1$ and $s_2$ to the nearest POI ($p_1$ and $p_2$). Based on the same POI database, different people's location histories become comparable. As depicted in Figure 20 B)-②, we can map a vehicle's trajectory onto a given road network using map matching algorithms, representing the trajectory as a sequence of road segment IDs, i.e. $r_1 \rightarrow r_2 \rightarrow r_3$. After that, we can compute traffic conditions on road segments based on trajectories mapped to them. As shown in Figure 20 B)-③, we can project trajectories of different moving objects, such as people, vehicles and animals, into uniform grids or regions with irregular shapes, counting in-and-out flows in each unit. We can also build links between two units if there are trajectories traversing them consecutively, constructing a transitional graph between locations.

As depicted in Figure 20 C), we can map entity names detected from different documents to the corresponding entity label, e.g. $e_1$ and $e_2$, and extract entity relations, like $l_1$ and $l_2$, using some classification methods. Likewise, we can match faces detected from different images against each other to find the same person across images.

### *7.3.3 Coarse Data Transformation*

Instead of detecting exact knowledge and matching them precisely, the coarse data transformation employs embedding algorithms or encoders to turn data into an intermediate representation which carries implicitly compressed knowledge.

1) Extensive studies on representation learning techniques have been done for compressing unstructured data into a dense and fixed length representation, which is further used in downstream applications. For example, different types of methods, including topic models, word embedding algorithms and autoencoders etc., were proposed for estimating continuous representations of texts.

- Topic models, like Latent Semantic Analysis (LSA) and LDA [6], represent a text document with a distribution of topics, each of which is further represented by a distribution of words.
- Autoencoders [3] consist of an encoder and a decoder, generating a compressed representation of the input data based on the bottleneck layer. The encoder compresses the input data into a lower dimensional representation, while the decoder reconstructs the original input from the compressed representation.
- Word embedding algorithms [2], such as neural network language models (NNLM) [5], word2vector [33] and BERT [13], learn a representation for each word using deep learning or matrix factorization, capturing the semantic relations between words.

In the meantime, different image encoders have been proposed to generate a latent space representation for tasks of image classifications, clustering and retrievals. Traditional PCA and recent convolutional autoencoders [31] are employed as an unsupervised representation learning model for images.

2) Quite a few research works employed embedding algorithms or encoders mentioned above to learn representations for spatio-temporal data [21][49]. Though those models reduce the computational complexity for downstream applications to some extent, the unique spatial and temporal properties introduced in Figure 17 are not well preserved in the latent representations. For example, as illustrated in Figure 17 A), location $s_1$ has a short distance to $s_2$ than $s_3$ in the geospatial space, i.e. $d_1 < d_2$. However, the distance between their latent representations generated by existing embedding models does not hold this, let along, the triangle inequality $d_2 - d_1 < d_3 < d_1 + d_2$ . That is, the spatial closeness and triangle inequality properties do not hold any longer in the latent representation space. This significantly reduces the key information contained in original spatio-temporal data, compromising the capabilities of downstream models and applications. Dedicated embedding algorithms or encoders, preserving unique spatial and temporal properties, should be designed for spatio-temporal data.

3) Regarding structured data, a wide range of network embedding algorithms [12] have been proposed to represent each node in a network with a lower-dimensional and dense vector that preserves the topology of network and/or content of the node. Networks can denote the social relations between friends, information relays between authors and biological interactions between molecules. However, there is lack of representation learning algorithms for embedding the subtitles in an online form, such as name, age, agenda, and address, which will be field names in databases. Those subtitles are an underlying graph of entities and their attributes rather than a sequence of words.

## 8. FUTURE DIRECTIONS

**1) Broaden and deepen the philosophy of knowledge alignment**: There may be more knowledge alignment principles except for the four proposed in this paper. In addition, it deserves a deep dive into each knowledge alignment principle, evaluating the strength of identified links between data.

**2) The combination of the two knowledge fusion paradigms**: As the two paradigms have their own advantages and disadvantages, combining them in an AI task may lead to a higher performance in some scenarios. For example, [14] has shown that combing large language models with knowledge graphs achieves a better result in linguistic tasks. Designing intelligent methods that can automatically generate precise knowledge fusion models is also an important research topic. For instance, [71] proposes a human-machine collaborative methods for generating precise knowledge graphs automatically.

**3) Dedicated data transformation models for different data modalities**: It is imperative to exploit the unique properties of each data modality, which should be well preserved in their latent representations. Thus, dedicated data encoders should be created for different data modalities and different categories of applications, capturing both unique properties of data and features of applications.

**4) Improving the efficiency and robustness of knowledge fusion**: As problems we are going to solve in the physical world usually request immediate responses with a high accuracy all day, the efficiency and robustness of knowledge fusion model matter. These issues could be handled from the perspective of data transformation, e.g. incorporating data compression and representation learning techniques that can significantly scaled down the size of input while maintaining the knowledge of row data.

**5) A deeper theoretical analysis on this topic**: There are still some questions remain open, needing a deeper theoretical analysis. For example, under what conditions does a cross-domain multimodal data fusion paradigm outperform single-domain models. Currently, we only have a general insight, i.e. when data in a target domain is insufficient to reflect the root causes of the problem to be solve, additional datasets from other domain should be included in a multimodal data fusion model. Or, can we derive generalization bounds or domain adaptation guarantees for the knowledge alignment principles.

## 9. CONCLUSION

In this paper, we exploit cross-domain multimodal data fusion systematically. A four-layer framework, containing ten procedures and three key components, is proposed for designing knowledge fusion methods to solve real-world problems. The first four procedures select useful datasets from a diversity of sources and domains, while the rest of procedures analyze and process data gradually. The first key component is the philosophy of knowledge alignment which reveals the nature of complementation between knowledge from different data regardless structures of AI models. With this component, we can discover the underlying links between disparate data and thus answer the question of “why can be fused”. The second key component is two paradigms of knowledge fusion, consisting of a precise fusion and a coarse fusion, which unveils the intrinsic differences between existing methods. The first and second components guide the direction of model selection. The third one is data transformation which turns data of different structures, resolutions, scales and distributions into a consistent representation. The second and third components lead to specific model structures, answering the question of “how to fuse”.

This paper is neither a research paper proposing a specifical algorithm to solve a particular problem nor a review on a broad range of research works that have been done. This paper points out a new research theme on data fusion from a new perspective (i.e. cross-domain, multimodality, and solving problems in the physical world). The paper paves the way towards this direction with a systematic framework, detailed procedures and concrete methodologies. This paper will not only facilitate problem solving in applications interacting with both digital and physical worlds but also inspire continued innovations in AI related research.

## ACKNOWLEDGMENT

This work is supported by the National Natural Science Foundation of China (62076191).

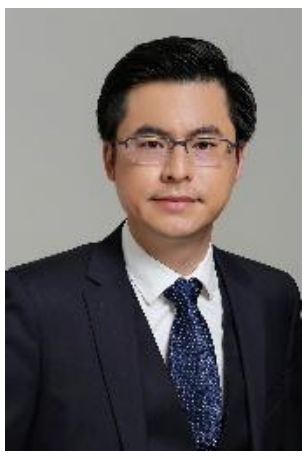

Dr. Yu Zheng is the Vice President of JD.COM and president of JD Intelligent Cities Research. Before Joining JD.COM, he was a senior research manager at Microsoft Research. He is also a chair professor at Shanghai Jiao Tong University. He was the Editor-in-Chief of ACM Transactions on Intelligent Systems and Technology (2016-2021) and has served as the program co-chair of ICDE 2014 and CIKM 2017. He is a keynote speaker of AAAI 2019, KDD 2019 Plenary Keynote Panel and IJCAI 2019 Industrial Days. He received SIGKDD Test-of-Time Award twice (in 2023 and 2024) and SIGSPATIAL 10-Year-Impact Award four times (in 2019, 2020, 2022, and 2023). He was named one of the Top Innovators under 35 by MIT Technology Review (TR35), an ACM Distinguished Scientist and an IEEE Fellow, for his contributions to spatio-temporal data mining and urban computing.